\documentclass[letterpaper]{article} %
\usepackage{aaai25}  %
\usepackage{times}  %
\usepackage{helvet}  %
\usepackage{courier}  %
\usepackage[hyphens]{url}  %
\usepackage{graphicx} %
\usepackage{natbib}  %
\usepackage{caption} %
\usepackage{algorithm}

\usepackage{newfloat}
\usepackage{listings}
\DeclareCaptionStyle{ruled}{labelfont=normalfont,labelsep=colon,strut=off} %
\floatstyle{ruled}
\newfloat{listing}{tb}{lst}{}
\floatname{listing}{Listing}
\title{Explaining Decisions of Agents in Mixed-Motive Games}
\author {
Maayan Orner\textsuperscript{\rm 1}, Oleg Maksimov\textsuperscript{\rm 1}, Akiva Kleinerman\textsuperscript{\rm 1}, Charles Ortiz\textsuperscript{\rm 2}, Sarit Kraus\textsuperscript{\rm 1}
}
\affiliations {
    \textsuperscript{\rm 1}Department of Computer Science, Bar-Ilan University, Israel\\
    \textsuperscript{\rm 2}SRI International, USA\\
    maayanorner@gmail.com
}

\usepackage{amsmath}
\usepackage{siunitx}
\usepackage{booktabs}
\usepackage{amssymb}
\usepackage[noend]{algpseudocode}
\usepackage[framemethod=TikZ]{mdframed}

\usepackage{enumitem}

\usepackage{xcolor}

\definecolor{austriared}{cmyk}{0, 0.9, 0.9, 0.3}  %
\definecolor{turkeyyellow}{cmyk}{0, 0.3, 1, 0.3}  %
\definecolor{italygreen}{cmyk}{0.7, 0, 0.8, 0.3}  %

\newcommand{\noparagraph}[1]{\textbf{#1}}

\begin{document}
\maketitle

\begin{abstract}
In recent years, agents have become capable of communicating seamlessly via natural language and navigating in environments that involve cooperation and competition, a fact that can introduce social dilemmas. Due to the interleaving of cooperation and competition, understanding agents' decision-making in such environments is challenging, and humans can benefit from obtaining explanations. However, such environments and scenarios have rarely been explored in the context of explainable AI. While some explanation methods for cooperative environments can be applied in mixed-motive setups, they do not address inter-agent competition, cheap-talk, or implicit communication by actions. In this work, we design explanation methods to address these issues. Then, we proceed to establish generality and demonstrate the applicability of the methods to three games with vastly different properties.
Lastly, we demonstrate the effectiveness and usefulness of the methods for humans in two mixed-motive games. The first is a challenging 7-player game called no-press Diplomacy. The second is a 3-player game inspired by the prisoner's dilemma, featuring communication in natural language. 
\end{abstract}

\section{Introduction}
\label{section:introduction}
Many important real-world scenarios resemble mixed-motive games. In these settings, agents' interests can be partly aligned or opposed, resulting in varying motivations to cooperate and compete.
In recent years, the study of automated agents that act in mixed-motive games has gained renewed attention within the community, using a 7-player game called Diplomacy as a research testbed \cite{paquette2019no,peskov2020takes,anthony2020learning,meta2022human,wongkamjan2024more}.

Meanwhile, the explainability of AI systems has been extensively studied \cite{Simonyan2013DeepIC, ribeiro2016should,lundberg2017unified,DBLP:conf/ijcai/Schleibaum0KM24}. Moreover, the focus on explainability has also expanded to various types of multi-agent systems \cite{ciatto2019towards,kraus2020ai,boggess2022toward,qing2022survey,guo2023explainable}. 
However, the unique aspects of explaining agents’ decisions in mixed-motive games have not been adequately addressed.

In mixed-motive games and analogous real-world situations \cite{GNYAWALI2011650}, although agents may possess conflicting interests \cite{gallo1965cooperative}, there is a competitive advantage for some cooperation \cite{kraus1994negotiation}. 
Furthermore, in mixed-motive games involving more than two agents, the decision of whom to cooperate with or compete against is crucial.
In these scenarios, agents' tendencies, the alignment of their interests, the actions they take, and the way they communicate shape cooperation patterns, which in turn affect the payoffs they receive \cite{parkhe1993strategic,sally1995conversation}.

To tackle these unique challenges, we define a framework of \emph{three conceptual levels} that explanations should address:
\begin{enumerate}
\item \textbf{Strategic}: The utility the agent obtains depends on its decisions, but also the decisions of other agents.
\item \textbf{Situational}:
The state of the environment and the policies of the participating agents can motivate cooperation or competition among them.
\item \textbf{Diplomatic:} Agents' actions convey information to other agents, which can influence future outcomes.
\end{enumerate}

Based on that, our research question is as follows:

\begin{mdframed}[
    linecolor=black, 
    linewidth=1pt,
    roundcorner=5pt,
    backgroundcolor=gray!10,
    innertopmargin=10pt,
    innerbottommargin=10pt,
    innerrightmargin=10pt,
    innerleftmargin=10pt,
    skipabove=9pt,
    skipbelow=9pt
]
\emph{\textbf{Main Research Question:}} Can we design explanation methods that address the strategic, situational, and diplomatic levels in mixed-motive games?
\end{mdframed}

The main contribution of this work is the development of methods for explaining agents' decisions in mixed-motive games, addressing the above research question.

We apply the methods in three environments with vastly different properties (see section \ref{section:envs} and appendix). To further demonstrate the applicability of the methods, agents in each environment employ a different type of policy: neural network in no-press Diplomacy, black-box (GPT-4) in Communicate Out of Prison (COP), and heuristic policy in Risk \cite{zuckerman2009mixing}.

Furthermore, we show the usefulness of the methods for humans through two user studies: the first in no-press Diplomacy and the second in COP, a 3-player game inspired by the prisoner's dilemma, featuring communication in natural language.

To the best of our knowledge, this is the first work to propose explanation methods specifically designed for mixed-motive games.

\section{Related Work}
During the last few years, researchers have proposed a new research area called Explainable Decisions in Multi-Agent Environments  
\cite{kraus2020ai}.
Various aspects of the topic have begun to be explored \cite{boggess2022toward,qing2022survey,guo2023explainable, zahedi2023didn}. 
However, 
most of that work has focused only on explaining agent interactions in cooperative settings \cite{nizri2022explaining,heuillet2022collective,yang2022design,mahjoub2023efficiently,angelotti2023towards}. 

Other studies have included experimentation in mixed-motive games, but these efforts remain partial.
For example, one study applied existing explanation methods \cite{khlifi2023diagnostics} in a mixed-motive game that focuses on coordination (Level-Based Foraging), while the other applied novel methods designed for cooperative setups to the same game \cite{boggess2023explainable}. A different study, which also focuses on coordination, does not evaluate the explanations relative to humans \cite{milani2022maviper}.
None of these studies aims to address the unique challenges we discussed in section \ref{section:introduction}.

Although solution concepts from cooperative game theory, such as Shapley values \cite{heuillet2022collective} or Myerson values \cite{angelotti2023towards} can provide a framework to examine the contribution of each agent to a cooperating team, they are insufficient for games with mixed motives. In such games, payoffs are typically not collective. Moreover, agents are not necessarily part of a team; they cooperate when beneficial but may become adversaries when motivated to compete. 

Explanations of feature attribution \cite{ribeiro2016should,lundberg2017unified} often contain many irrelevant details while lacking relevant information. For example, in Diplomacy, such methods can be utilized to estimate the contribution of each unit movement (see section \ref{subsection:no_press_diplomacy}) to the utility value of a strategy (see section \ref{subsection:preliminary_def}). Based on preliminary experimentation, such explanations are difficult to interpret. More importantly, they do not address questions related to agents' interactions. Therefore, new explanation methods are needed.

We drew a considerable amount of inspiration and designed the estimation algorithm for Strategy-based Utility Explanations (section \ref{subsection:sbue}, figure \ref{fig:diplomacy_explanation} upper heatmap) based on the Simulation Action Value Estimation (SAVE) algorithm from \cite{kramar2022negotiation}. Shared Interests Correlation Analysis (section \ref{subsection:sica}, figure \ref{fig:diplomacy_explanation} lower heatmap), although developed independently, is similar to a technique utilized by  \cite{zhang2021coordination} for a different task.
Presenting the probable actions other agents might take (section \ref{likely_actions}, figure \ref{fig:diplomacy_explanation} arrows) is a special case of example-based explanations \cite{cai2019effects}.

Large language model (LLM) agents are capable of acting in mixed-motive games \cite{meta2022human}. To further examine the diplomatic level, we apply our explanation methods in setups with communication; it requires the development of LLM agents' game simulations, a topic that has received considerable attention \cite{yan2023larp, akata2023playing,xu2023exploring,mukobi2023welfare,gemp2024states}.

\section{Environments}
\label{section:envs}
To examine different explanation methods, we explored alternative game settings. The first was no-press Diplomacy, a mixed-motive game with a large action space that introduces \emph{situations} that encourage cooperation or competition between agents. In addition to tactical and \emph{strategic} considerations, the game requires \emph{diplomatic} understanding, making it suitable for studying our research question.

Since research has shown that cheap-talk can alter cooperation rates in games similar to the prisoner's dilemma \cite{sally1995conversation}, we designed the Communicate Out of Prison (COP) game. COP is strategically simple to minimize confounding factors but features communication in unconstrained natural language.
The LLM agents we developed for COP can adopt different personality types. This allows us to examine the explanations in settings where heterogeneous agent policies lead to different cooperation patterns. COP is defined as a 3-player game, to introduce rich social dynamics.

To broaden the scope of our evaluation, we also applied the methods in a simplified version of Risk (see appendix).

\subsection{No-Press Diplomacy}
\label{subsection:no_press_diplomacy}
Diplomacy is a simultaneous game in which each player controls one of the seven great powers of Europe in the years leading to World War 1. The goal of the game is to control at least 18 out of the 36 strategic locations on the map (``supply centers"). An action in Diplomacy is composed of multiple unit sub-actions (e.g., Naples fleet to Ionian Sea, Rome army to Apulia,...), yielding a large combinatorial action space. The game is estimated to have $10^{21}$ - $10^{64}$ joint actions per turn and a game tree size that can be infinitely large (median size $\approx 10^{896.8}$) \cite{anthony2020learning}.

There are two notably popular versions of Diplomacy, one permits explicit communication (full-press) \cite{kraus1995designing}, while the other relies on implicit communication through in-game actions (no-press).

For our experiments, we use the game environment from \cite{paquette2019no}, along with a neural policy network and value function from \cite{anthony2020learning}.

\subsection{Communicate Out of Prison Game}
\label{COP_and_Risk}
In this game, which draws inspiration from the prisoner's dilemma and the pirate game \cite{azaria2015agent}, three agents 
(denoted as $\{a,b,c\}$) attempt to avoid punishment for a robbery.
The game starts with a communication stage, in which agents exchange private messages sequentially. 
After the communication stage is over, 
every agent announces whether each of the other agents is innocent or guilty; all agents announce simultaneously. For example, agent $a$ can announce $(b=guilty$, $c=innocent)$. 
The payoffs, determined by the announcements, are designed to motivate both cooperation and competition.

For our experiments, we defined three types (private information) and prompted the LLM agents to play accordingly:
\begin{itemize}
  \item \textbf{con-artist:} cruel, manipulative, and deceitful.
  \item \textbf{``simple-person":} nice, trusting, honest, and hates lies.
  \item \textbf{politician:} a political genius, selfish but rather honest, prefers ``simple and nice" agents, dislikes manipulators.
\end{itemize}

\section{Explanation Methods}

\label{section:methods}
\begin{figure}[t]
\centering
\includegraphics[width=1.0\columnwidth]{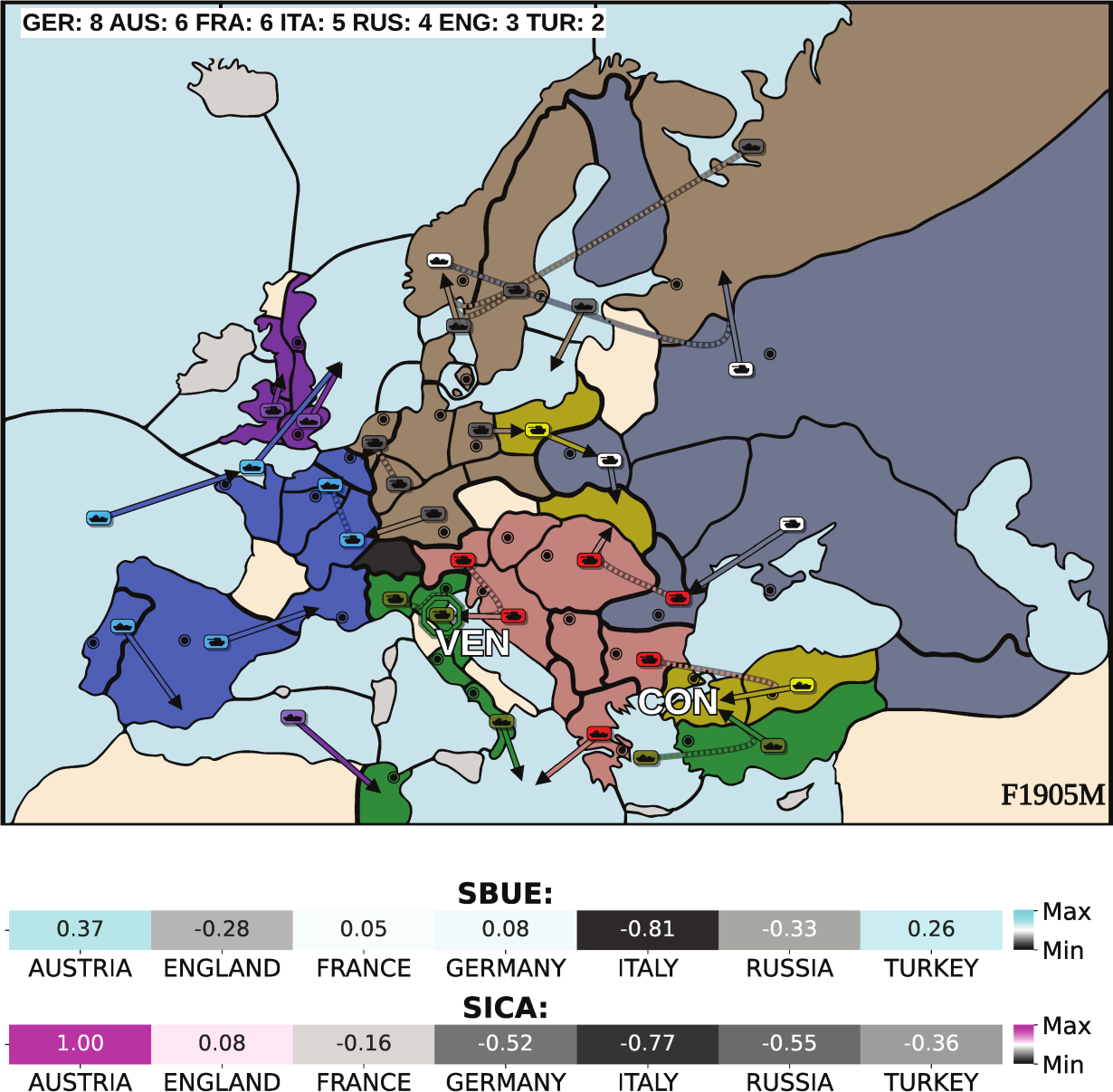}
\caption{Explanation for \textbf{{\color{austriared}Austria's}} strategy $a^i$, where it assists \textbf{{\color{turkeyyellow}Turkey}} in preventing \textbf{{\color{italygreen}Italy}} from taking over Constantinople while attacking Venice. \emph{SICA} detects animosity with Italy; \emph{SBUE} explains that $a^i$ implicitly communicates hostility to Italy and friendliness to Turkey. Austria's arrows visualize $a^i$; arrows of others present their \emph{probable actions}.}
\label{fig:diplomacy_explanation}
\end{figure}
In this section, we present the explanation methods. In the appendix, we provide additional pseudo-code and a method to find counterfactual actions in Diplomacy.
Assuming we play as agent $i$, and would like to get an explanation for action $a^i$ in state $s$, we propose three different explanation methods:
\textbf{(1)} A utility-based method that presents the expected utility associated with $a^i$ both for agent $i$ and the other agents in the environment (section \ref{subsection:sbue}). \textbf{(2)} A probable actions-based method that specifies the most probable actions other agents can take when agent $i$ performs $a^i$ (section \ref{likely_actions}). \textbf{(3)} An estimation of the relationships between the agents in state $s$ (section \ref{subsection:sica}). A visual example of all three types of explanations in Diplomacy is provided in figure \ref{fig:diplomacy_explanation}.
\subsection{Preliminary Definitions}
\label{subsection:preliminary_def}
\noparagraph{Environment and agents:} An environment of $p$  agents incorporates an action space $A$, state space $S$, transition function ${T:S \times A^{p} \rightarrow S}$, and reward function ${R: S \times S \rightarrow \mathbb{R}^{p}}$. Considering the current state as $s_t$ (the subscript is occasionally omitted), we denote $R(s_{t-1},s_{t})$ as $R_{t}$, and the corresponding reward for agent $i$ as $R_t^i$. 

The set $P$ consists of agents that act according to policies $(\pi_1,...,\pi_p)$. A policy ${\pi_i: S \times A \rightarrow [0,1]}$ is a joint probability mass function.
For each agent, we define a value function ${V_i:S \rightarrow \mathbb{R}}$, which estimates the expected return of the agent given $\pi$; i.e. ${V_i(s_t) \approx \mathbb{E}[R^i_{t+1}+...+\gamma^{n-1}R^i_{n}|s_t]}$, where ${\gamma \in [0,1]}$ is a discount factor.
For convenience, we define ${V: S \rightarrow \mathbb{R}^{p}}$ as a function that applies $(V_1,...,V_p)$, and ${\pi: S \times A^{p} \rightarrow [0,1]^{p}}$ as a function that returns the probability of agents' actions according to $(\pi_1,..,\pi_p)$.
We denote $a^i \sim \pi_i$ as drawing an action from the distribution $\pi_i$ given state $s$ ($s$ is implied), and $a \sim \pi$ as a vectorized version of it. We use $-i$ to denote all agents that are not $i$.

\noparagraph{Utility of an outcome:} Starting from state $s_t$, the utility vector of a specific outcome is defined as $\gamma V(s_{t+1})+R_{t+1}$.

\noparagraph{Expected utility of an action:} The expected utility vector of action ${a^i \in A}$ is a vector of size $p$, in which the element with index $j$ corresponds to the expected utility of action $a^i$ for agent $j$.
Formally, it is defined as ${\mathbb{E}_{a^{-i} \sim \pi_{-i}}[\gamma V(T(s,(a^i,a^{-i})))+R_{t+1}]}$, where $\pi_{-i}$ are the policies of agents $-i$ or the policies we assume they employ.

\begin{algorithm}[tb]
\small
\caption{Simulate}
\label{algorithm:simulation}
\begin{algorithmic}[1]
    \Procedure{Simulate}{$s \in S$, $k \in \mathbb{N}$, $d \in \mathbb{N}$, $C$}
        \State // $C$ is a set of tuples $(a^i_e,d_e)$ where $a^i_e \in A$ and $d_e \in \mathbb{N}$
        
        \State $X \gets k \cdot d \times p$ matrix; $s_{0} \gets s$
        \For{$j \gets 0$ to $k-1$} // K simulations
            \State $s \gets s_0$ // restore initial state
            \State $r_{cumulative} \gets <0,...,0>$ // vector of size $p$
            \For{$t \gets 0$ to $d-1$} // to depth d
                \State $a \gets a^1,..,a^{p} \sim \pi$ // draw actions for all agents
                \For {$(a_e^i, d_e) \in C$}
                    \If{$t = d_e$}
                    $a^i \gets a_e^i$   // replace with $a_e^i$.

                    \EndIf
                \EndFor
                \State $s' \gets T(s,a)$
                \State $r_{cumulative} \gets r_{cumulative}+\gamma^{t} R(s,s')$
                \State utility $\gets \gamma^{t+1} V(s') + r_{cumulative}$
                \label{algorithm:simulation:utility}
                \State $X[j \cdot d+t, :] \gets$ utility; $s \gets s'$
            \EndFor
        \EndFor
        \State \textbf{return} $X$
    \EndProcedure
\end{algorithmic}
\end{algorithm}

\subsection{Strategy-Based Utility Explanations (SBUE)}
\label{subsection:sbue}
\noparagraph{Motivation and description:} 
When engaging in mixed-motive games with more than two agents, it is advantageous to consider not only the action's benefit to the agent itself (\emph{strategic}) but also its influence on other agents observing this action (\emph{diplomatic}). This is crucial due to the role of implicit communication in these games. For example, a ``friendly" action can communicate willingness for future cooperation, whereas a ``hostile" action usually does not.

SBUE addresses both the strategic and diplomatic levels by explaining how an action influences the game outcomes for all agents, \emph{presenting the expected utility value for each agent} conditioned on that action.

\noparagraph{Explanation estimation:}
To explain action $a_e^i$ ($e$ denotes explained) given state $s$, the following steps are performed:
\begin{enumerate}
\item Simulate the next turn from $s$ for $k$ times, where agent $i$ performs action $a_e^i$, and all other agents follow their respective policies.
\item Estimate the utility values of each outcome using the value functions and rewards (algorithm \ref{algorithm:simulation} line \ref{algorithm:simulation:utility}).
\item Estimate the expected utilities of $a_e^i$ by computing the mean utility of each agent (column); return a vector of size $p$.
\end{enumerate}
Steps 1 and 2 are equivalent to using algorithm \ref{algorithm:simulation}:
\[
{Simulate(s,k,d=1,C=\{(a_e^i, 0)\})}
\]

In cases where the value function is difficult to interpret, we estimate $\mu^i$ and $\sigma^i$ for all $i \in P$ via unconstrained simulation (algorithm \ref{algorithm:simulation} with $C=\emptyset$), and perform Z-score standardization to each column before step 3. 

\subsection{Probable Actions-Based Explanations}
\label{likely_actions}
\noparagraph{Motivation and description:} In any environment with more than two agents, understanding the policies of the other agents can be useful. Therefore, we present the most probable actions of agents $-i$, assuming agent $i$ selects action $a_e^i$.
This explanation was (primarily) designed to address the \emph{strategic} level. It can be viewed as an example-based explanation, as it presents an example of a probable outcome assuming the agent plays $a_e^i$.

\noparagraph{Explanation estimation:} As in SBUE, we run $k$ simulations from state $s_t$, where agent $i$ performs action $a_e^i$ and all other agents follow their respective policies. Then, we extract the most commonly used action of each agent accordingly. This explanation can be extended to multi-turn trajectories by greedily repeating the process where all agents, including $i$, follow their respective policies.
In our Diplomacy user study, we present one turn since understanding longer trajectories is cognitively challenging (see figure \ref{fig:diplomacy_explanation}, arrows).

\noparagraph{Limitations for LLM agents:} Finding the most commonly used actions of LLM agents is challenging because of the complexity of natural language. As a workaround, we decode each action (by decoding tokens) using temperature ${\tau = 0}$. 
While this solution is consistent with our greedy approach, it can lead to incorrect explanations (see section \ref{section:llm_evaluation_summary}).

\subsection{Shared Interests Correlation Analysis (SICA)}
\label{subsection:sica}
\noparagraph{Motivation and description:} In mixed-motive setups, a central question is whether the state of the environment and agents' policies can facilitate effective cooperation. 
To explain the cooperation tendencies and alignment of interests among pairs of agents in a given state, we introduce Shared Interests Correlation Analysis (SICA).

The SICA value of agents $i$, $j$ is the Pearson correlation coefficient of their obtained utilities. When causal relationships can be assumed (i.e., the actions of $i$ can influence the payoff of $j$), the SICA value can be interpreted as a measure of the ``friendliness" or cooperativeness of $i$ and $j$.
 
SICA is typically accompanied by an action-based explanation. This pairing reveals the impacts of actions on friends and enemies, presenting a fuller diplomatic picture. For example, it highlights how hostile actions toward
friends differ from those toward enemies, addressing the \emph{situational} and \emph{diplomatic} levels in our framework.

\noparagraph{Explanation estimation:} To estimate SICA, we perform $k$ unconstrained simulations to depth $d$ according to algorithm \ref{algorithm:simulation}, where all agents act according to their respective policies, i.e., $a \sim \pi$. This step is equivalent to calling:
\[
{X \leftarrow Simulate(s,k,d,C=\emptyset)}
\]
The resulting dataset $X$, considering the case of depth-1 ($d=1$) without loss of generality, is a $k \times p$ matrix, in which element $X_{x,y}$ is the utility agent $y$ obtains in simulation $x$ (algorithm \ref{algorithm:simulation} line \ref{algorithm:simulation:utility}). Lastly, we compute the sample Pearson correlation coefficient for each pair of columns (i.e., agents), which results in a $p \times p$ correlation matrix.

\noparagraph{Interpretation of SICA:} The correlation coefficients are determined by the environment and policies of the agents, which we refer to as \emph{conditions}.
A \emph{strong positive correlation coefficient} between agents $i$ and $j$ indicates that the conditions lead to mutual benefit and cooperation. 
A \emph{strong negative correlation coefficient} suggests that the conditions lead to conflicting interests, resulting in competition or antagonism. 
A \emph{weak correlation coefficient} (near zero) may result from conditions that lead to inconsistent impacts of one agent's actions on another, reflecting interactions that range from indifference to complex, nonlinear patterns.
 
\noparagraph{Interdependence theory:} The definition of SICA is analogous to the concept of covariation of interests \cite{van2015interdependence}, drawing a connection to interdependence theory \cite{kelley1959social,kelley1978interpersonal}.

\section{Modules Evaluation}
This section presents the results of our evaluation experiments in Diplomacy and Risk. The experiments conducted in the COP game are summarized here and described in detail in the appendix.

\subsection{Convergence of Estimation of SICA and SBUE in Diplomacy}
\begin{figure}[t]
\centering
\includegraphics[width=\columnwidth]{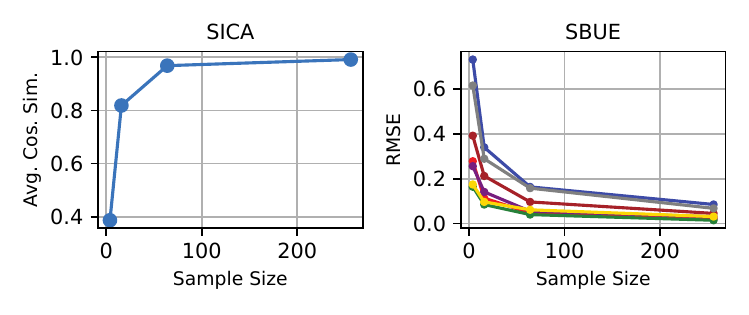}
\caption{Convergence of SICA and SBUE in Diplomacy. On the right side, each line corresponds to a different agent.}

\label{fig:ssize_plot}

\end{figure}
Since the run-time of both methods increases linearly with the number of samples and sampling is expensive, we examine the error-runtime trade-off for the methods using several sample sizes. For both methods, we perform 50 independent repeated estimations for each sample size $k$. 

For SBUE, we estimate once with $k=2,500$ and define it as ground truth. Then, we repeatedly estimate for each $k$. The RMSE (root mean square error) for each agent's value is computed independently relative to its corresponding ground truth. Small RMSE values indicate that $k$ is sufficient to provide reliable estimates.
For SICA, for each $k$, we repeatedly estimate and compute the average cosine similarity among all pairs of (flattened) correlation matrices. When the average cosine similarity is high, it implies convergence.

In Diplomacy, although the action space is large, we observe that SICA and SBUE require reasonable sample sizes in the game states we examined, as shown in figure \ref{fig:ssize_plot} (plot of typical results in a middle-game state). However, the convergence rates are highly dependent on the specific setup.

\subsection{Evaluation of SICA in Diplomacy and Risk}
We hypothesize that \emph{SICA's estimation of the relationships between agents is well-aligned with humans' opinions}, as long as the agents play similarly to humans.
\paragraph{Setup:}
\label{subsection:diplomacy_risk_sica_eval}
To test our hypothesis, 
we conducted a human-based experiment, in which we asked human players to annotate the most friendly and hostile agents in multiple board states, assuming they play one of the roles (e.g., Austria).

Specifically, we randomly generated 30 Diplomacy game states (ensuring representation from different game stages) and asked human players to annotate the top two hostile and friendly agents for each board. If an annotator was unable to decide the identity of the second most friendly or hostile agents, we allowed the determination to be left unfilled.  For Risk (4-player version), we generated 12 board states, and a human selected the top enemy and top friend for each board. 

We used two different annotators for diplomacy: one was considered a strong player, and the other was an intermediate player (introducing a variation of skill level). For Risk, which is not our main focus, the dataset was annotated by the authors.

\noparagraph{Metric and evaluation:} To evaluate the alignment of the annotations with SICA, we used MAP@K \cite{manning2009introduction}. We ranked the other agents by the SICA value they shared with the agent, high to low, and reversed the ranking to rank agents by hostility. Then, we evaluated SICA using the annotated datasets. This methodology was built upon the assumption that agents play similarly to humans, and a violation of it is likely to result in a decrease in the MAP@K values.

\noparagraph{Inter annotator agreement (IAA):} To compare SICA to the agreement between annotators, we computed lower and upper bounds for MAP@K for the annotations of $A$ compared to the partial ranking of annotator $B$ (called IAA-rank in table \ref{table:map_score}). Note, ${MAP@K(A,B_{rank}) \neq MAP@K(B,A_{rank})}$, but we do not expect symmetry in this setup.

\noparagraph{Results:}
The results (see table \ref{table:map_score}) suggest that SICA is well-aligned with human intuition and performs better than random rankings in both Diplomacy and Risk. In Diplomacy, SICA outperforms a heuristic-based ranking, which sorts agents by the number of centers they own (a proxy for strength), by a considerable margin. This is achieved without relying on non-generalizable properties of the environment.
To clarify, the results for the random rankings of friends (0.58) and enemies (0.55) differ because the expert, as allowed, selected different numbers of friends and enemies.

The range of IAA-rank was higher than SICA in 2 out of 4 cases, lower in one case, and overlapped in another. In other words, in this setup, the degree of alignment between SICA and humans is comparable to the degree of agreement among humans themselves.

\begin{table}[tb]
\small
\centering

\begin{tabular}{c c c c | c c}
\toprule
 Category &
 Rand. & Cen. & SICA & IAA-rank & N \\ 
\midrule
(D, e) E  &0.55 & 0.63&\textbf{0.74} & $0.86 \pm 0.01$ & 30\\
(D, e) F &0.58&0.60&\textbf{0.66} & $0.725 \pm 0.015$ & 30 \\
\hline
(D, p) E  &0.56 &0.74&\textbf{0.81} & $0.82 \pm 0.01$ & 30\\
(D, p) F &0.56& 0.58&\textbf{0.78} & $0.72 \pm 0.02$ & 30 \\
\hline
(R) E &0.61&-&\textbf{0.81}&-&12\\
(R) F &0.61&-&\textbf{0.72}&-&12\\
\bottomrule
\end{tabular}
\caption{
MAP@K score, Diplomacy (D, e) - expert annotation, (D,p) - non-expert annotation, Risk (R); E - enemies, F - friends; Rand. - Random, Cen. - Centers.}

\label{table:map_score}
\end{table}

\subsection{LLM Evaluation Experiments (COP)}
\label{section:llm_evaluation_summary}
\noparagraph{Types consistency: }First, to validate that LLM agents mimic the types we defined consistently, we asked two humans to detect the type of each agent in anonymized games (based on communication only; payoffs were hidden). We observed that humans can distinguish between the types (accuracy: annotator 1 -  86.66\%; annotator 2 - 93.33\%).

\noparagraph{Preliminary experiments:} We conducted experiments to examine the methods in a setup with LLM agents.

\textbf{(1)} We defined the agents' personalities to encourage cooperation between the politician and the ``simple-person" against the con-artist (see section \ref{COP_and_Risk}). Therefore, we hypothesized that SICA would explain it.

First, we generated the explanation in a standard setup with a politician, a ``simple-person", and a con-artist. The explanation was aligned with our hypothesis. To compare that explanation with another explanation and to ensure that SICA would reflect changes in the agents' dynamics, we repeated the process in a different setup involving two politicians and one ``simple-person".
In this case, as we expected, the explanation showed balanced cooperation and competition patterns among all the agents.

\textbf{(2)} To examine SBUE, we manually curated two messages from the politician to the ``simple-person": a friendly message ($m_f$) and a hostile message ($m_h$).
The SBUE explanations indicated that $m_f$ was more beneficial to both agents, while $m_h$ provided an advantage to the con-artist, which was consistent with our hypothesis.

\textbf{(3)} For the probable actions-based explanations, we examined how the temperature parameter affected the game outcomes. We found that using a temperature $\tau = 0$, which corresponds to greedy decoding (our approach), sometimes led to outcomes that were not probable when using $\tau = 0.7$. This highlights a significant limitation: the explanations produced using our (simplistic) approach can be misleading.

\noparagraph{COP large language model study:} Lastly, we studied how SICA and SBUE influence LLM agents' decisions.
This study mirrors (to a large extent) the study described in section \ref{section:llm_user_experiment}, but here the ``participants" are LLM agents instead of humans.
Our findings suggest that agents follow the explanations in most cases. We demonstrated that the explanations can help mitigate a well-known bias in LLMs, where decision-making is affected by the order of presented options \cite{wang2023large}. This highlights the potential of the explanations to enhance LLM agents' decision-making in the setups we study.

\section{User Studies}
We conducted two complementary studies with humans in two different environments. Our \emph{Diplomacy user study} includes a comparison of multiple explanation types and a general examination of the effectiveness of each explanation type with respect to our research question. 

The \emph{COP user study} was designed based on the results of the Diplomacy study. Here, we further examine the effectiveness of SICA and SBUE in a setup with explicit communication and heterogeneous policies — focusing on specific aspects of the situational and diplomatic levels.

\noparagraph{Questionnaire design:} In both experiments, users answered questionnaires where each response was on a Likert scale (1-5).
They were asked the same set of questions before and after being presented with an explanation, and we examined how the explanation modified their answers. 
We tested our hypotheses using the Wilcoxon signed-rank test with a $\alpha$ value defined as $0.05$. In the Diplomacy user study, we asked additional questions after presenting each explanation to assess how it was perceived (see section \ref{section:study_design}).

\subsection{Diplomacy User Study}
\label{section:study_design}
We recruited 26 subjects (24 males, 1 female, 1 other), with the requirement that players were familiar with the rules of Diplomacy.
Participants who provided inconsistent answers to our consistency check questions or failed a basic knowledge exam were discarded automatically (3 out of 26).

\noparagraph{We presented users with 3 types of explanations:}
\begin{figure}[tb]
\centering    \includegraphics[width=\columnwidth]{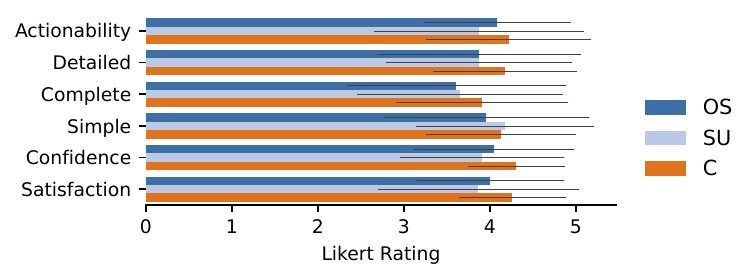} 
    \caption{Mean and SD of participant ratings of explanations. Participants almost consistently prefer explanation \emph{C}.}
    \label{fig:comparing}

\end{figure}
\begin{itemize}
    \item \emph{OS} (others' strategies): presents the probable actions of other agents (see section \ref{likely_actions}; figure \ref{fig:diplomacy_explanation} - arrows).
    \item \emph{SU} (shared interests and utilities): a combination of SICA and SBUE (sections \ref{subsection:sica}, \ref{subsection:sbue}; figure \ref{fig:diplomacy_explanation} - heatmaps).
    \item \emph{C} (combined): combines explanation types \emph{OS} and \emph{SU}.
\end{itemize}

\noparagraph{Experiment setup:} The experiment covers three different Diplomacy boards (states); in each of them, the user is presented with a strategy (unit movements) with higher expected utility together with a counterfactual. Each board is paired with a different explanation type, where each type is presented only once. We followed a round-robin scheme to determine \emph{(state, explanation)} allocations and temporal explanation type order, to preserve balance and prevent biases. 

\noparagraph{Hypotheses:} We make two hypotheses that cover the conceptual levels we defined:

\begin{itemize}
\item H1: The explanation method improves users' performance at the strategic level.
\item H2: The explanation method enhances users' understanding of the situational and diplomatic levels.
\end{itemize}

\emph{To test H1:}
Users were asked to assess their willingness to choose each strategy before and after seeing the explanation.

\emph{To test  H2:}
Users were asked to rank their understanding of the effects of their preferred strategy on ``friends and enemies" before and after seeing the explanation. This requires understanding which agents are considered either friends or enemies (\emph{situational}), and how the strategy influences agents within these identified groups (\emph{diplomatic}).

\noparagraph{Results:} 
As seen in table \ref{tab:proportions}, the effect of the explanations is statistically significant in all cases, except for H2 explanation \emph{OS}. For the \emph{strategic} level, H1($\pm$), the ratio of cases where users follow the explanations and raise (lower) their willingness to select the superior (inferior) strategy is at least as high as 43\% for all types (\emph{SU}, \emph{OS}, and \emph{C}).

Notably, the proportion of subjects who rated their understanding of agents' alignment of interests, and how a strategy influences other agents (H2), is nearly twice as high when SICA and SBUE are provided with or without example-based explanations, in comparison to example-based explanations alone (\emph{C}: $41\%$, \emph{SU}: $36\%$ vs \emph{OS}: $23\%$).

In all cases, combined explanations \emph{C} are at least as effective as explanations \emph{SU} or \emph{OS} individually, as shown by the bolded values in the table.

\begin{table}[tb]
\small
\centering
\begin{tabular}{cccccc}
\toprule
H & Exp. & Lower & Higher & p-val. & N-subj.\\
\midrule
H1 (+)&OS&0.17&0.57&0.005&23\\
&SU&0.09&\textbf{0.65}&0.001&23\\
&C&0.00&\textbf{0.65}&0.001&23\\
\hline
H1 (-)&OS&0.52&0.13&0.012&23\\
&SU&0.43&0.00&0.004&23\\
&C&\textbf{0.57}&0.13&0.007&23\\
\hline
H2 (+)&OS&0.05&0.23&0.102*&22\\
&SU&0.05&0.36&0.021&22\\
&C&0.05&\textbf{0.41}&0.012&22\\
\bottomrule

\end{tabular}
\caption{Diplomacy - the proportion of instances where users raised/lowered their answers after seeing the explanation. In H(+) we hypothesized an increase and in H(-) a decrease.}
\label{tab:proportions}

\end{table}

\noparagraph{Comparing explanations:}
In addition to the main part of the experiment, we defined a set of favorable properties based on previous work  \cite{hoffman2018metrics,boggess2022toward,boggess2023explainable}. For each explanation, we asked participants to rate these properties on a Likert scale.

Given the sample size, the statistical power was insufficient to detect significant differences per property. However, aligned with our intuition, \emph{explanation C (combined) is rated higher in every category except for simplicity} (6 out of 7, see figure \ref{fig:comparing}). That result further demonstrates the complementary nature of the methods and is consistent with the main results of the experiment (table \ref{tab:proportions}).

\subsection{Communicate Out of Prison User Study}
\label{section:llm_user_experiment}
We recruited 38 subjects (31 males, 7 females; all are graduate-level computer science students) who did not participate in the Diplomacy user study. Subjects who stated they did not understand the explanations were discarded automatically (5 out of 38). 

The focus of this study is to examine whether SICA and SBUE\footnote{The limitation we discuss in section \ref{likely_actions}, and the results of our Diplomacy user study motivated us to exclude the probable actions-based explanations from this study.} are useful for humans in environments where agents communicate via natural language. 
To evaluate this, we selected Communicate Out of Prison (COP) as a testbed. Recall that in COP, three
agents communicate with each other privately, trying to stay out of
prison by convincing others to side with them.

\noparagraph{Experiment setup:} We presented users with one to three dialog states in a randomized order. The number of states varied because users were allowed to finish the experiment early. 

To generate the states, we simulated the game until it included some chat history of the playing agent with both other agents, to enable users to form opinions about the other agents' personalities and tendencies. The agents played according to the personality types we defined in section \ref{COP_and_Risk}.

In each state, we presented users with an action (message to one of the other agents) with a higher expected utility and a counterfactual. For example, a possible action is \emph{``A to B: Listen here, mate. C's been tryin' to convince me it's you ... But I ain't buyin' it. You've been straight with me, ... We stick together ... Let's point at C."}

\noparagraph{Hypotheses:} We make three different hypotheses, one is related to the effectiveness of the explanation methods at the \emph{strategic} level, and the others refer to aspects of the \emph{diplomatic} and \emph{situational} levels:
\begin{itemize}
\item H3: The explanation method improves users' performance at the strategic level in mixed-motive games with explicit communication via natural language.
\item H4: The explanation method improves the understanding of inter-agent relationships (with respect to the chat history and the policies of the agents) in mixed-motive games with explicit communication via natural language.
\item H5: The explanation method improves the understanding of the diplomatic influence of explicit communication actions in mixed-motive games.
\end{itemize}

\emph{To test H3:} Users were asked which of two messages should be sent, before and after observing the explanation.

\emph{To test H4:} Users were asked which agent out of the other two was more friendly and trustworthy, before and after being presented with the explanation.

\emph{To test H5:} Users were asked to rate their understanding of the influence of the messages on the other agents, before and after being presented with the explanation.

\noparagraph{Results:} As seen in table \ref{tab:proportions_llms}, the explanations modified the opinions of humans significantly, both about the question of which message should be sent ($73\%$), as well as their understanding of inter-agent relationships ($59\%$). The explanations also subjectively improved users' understanding of the influence of each message on the other agents ($30\%$).

\begin{table}[t]
\small
\centering
\begin{tabular}{ccccc}
\toprule
H & Lower & Higher & p-val.  & N-subj.\\
\midrule
H3 (+)
&0.05&0.73&\num{3e-9}&33\\
\hline
H4 (+)
&0.09&0.59&\num{5e-7}&33\\
\hline
H5 (+)
&0.076&0.30&0.001&33\\
\bottomrule
\end{tabular}
\caption{COP - the proportion of instances where users
raised/lowered their answers after seeing the explanation. H(+) indicates we hypothesized the answer would be raised.}

\label{tab:proportions_llms}
\end{table}

\section{Discussion, Limitations, and Future Work}
\label{section:discussion}
\noparagraph{Discussion:} In this work, we presented methods to explain the decisions of agents that act in mixed-motive environments, focusing on games with more than two agents. First, we briefly discussed some of the challenges of mixed-motive games and described what explanation methods should address, using a three-level framework. Based on that, we designed explanation methods, including a method to explain relationships between agents in mixed-motive games.
We applied these methods in three environments, one of which involved LLM agents with distinct personality types.

We conducted two user studies to evaluate the usefulness of the methods for humans and found the explanations helpful. We observed that SICA and SBUE are effective for the tasks they were designed for but likely enhanced when combined with example-based explanations. We conclude that the proposed methods are complementary to some extent.

\noparagraph{Limitations:}
The main limitation of this work has to do with the fact that the subject is understudied, making a comparison with appropriate well-established baselines difficult.
Since our methods are simple to apply, we believe they can be adopted as baselines in mixed-motive games. 

In both user studies, participants were sampled from groups of potential users, resulting in gender imbalances that reflect the groups' demographics. For example, 85 out of 87 participants at the 2014 World DipCon were males \cite{Hill_2014}. Similarly, the COP study mirrors the male-to-female ratio in many subfields of computer science \cite{yamamoto2022gender}.

\noparagraph{Future work:} 
The proposed solutions are designed to be simple and general, inviting further extensions and improvements. For example, in SICA, any association function can be substituted for instead of Pearson's $r$.

Future work can also examine our (or novel) explanation methods for agents that act in mixed-motive games involving natural language and strategic elements, such as full-press Diplomacy. Additionally, developing LLM agents that act in simulated environments is an expanding research area (see related work), and investigating the explanation methods in complex LLM-based environments can be valuable.

\section*{Acknowledgments}
This research has been partially supported by the Israel Ministry of  Innovation, Science \& Technology grant  1001818511.

\appendix
\bibliography{refs}
\onecolumn
\section{Communicate Out of Prison (COP)}
\subsection{Environment}
\label{subsection:out_of_prison_game}
In this game, which draws inspiration from the prisoner's dilemma, three agents 
(denoted as $\{a,b,c\}$) attempt to avoid punishment for a robbery.

The game starts with a communication stage, in which agents exchange private messages sequentially. 
After the communication stage is over, 
every agent announces whether each of the other agents is innocent or guilty; all agents announce simultaneously. For example, agent $a$ can announce $(b=guilty$, $c=innocent)$. 
The payoffs, which are determined by the announcements, were designed to motivate both cooperation and competition.

To demonstrate why the game is cooperative-competitive - if every agent announces the two other agents as innocent, the payoffs will be $\{a:-5, b:-5, c:-5\}$. In case every agent announces the two other agents as guilty, the payoffs are $\{a:-20, b:-20, c:-20\}$. However, if two agents, e.g., $a$ and $b$ cooperate against agent $c$, and announce ${b=0, c=1}$ and ${a=0, c=1}$ respectively, the payoffs are $\{a:0,b:0,c:-20\}$.

\noparagraph{\emph{COP - Protocol of communication:}}
Agents communicate sequentially for $K$ ($K=4$ in our experiments) rounds in a randomized precedence order. In each round, each agent must send a message to one of the other agents.
During round ${K+1}$, the agents make announcements and obtain payoffs accordingly.
Communication is performed using unconstrained natural language, therefore, state space $S$ is all possible chat histories, and action space $A$ is all possible messages. $S$ and $A$ are unbounded in theory, but in practice, denoting $|C|$ as the maximal context length, and $|G|$ as the maximal message length, $|S|>2^{|C|}$ and $|A|>2^{|G|}$ (length = number of tokens).

\noparagraph{COP simulation components:}
Our simulation environment follows the communication protocol presented above and supports agents' augmentation with different personality types. Agents' policies are implemented using GPT-4 (instruct), by a prompt structured as \emph{\#setup\#meta-context(agent-id, task, personality, chat-history, stage)}. \emph{\#setup} refers to technical specifications such as the message structure. \emph{\#task} refers to the rules of the game, and \emph{\#personality} contains a detailed textual description of the personality type the agent is instructed to mimic. \emph{\#stage} is a specification of whether it is time to announce or to communicate.
Agent types are private and an agent is unaware of the possible types of other agents; the types can be estimated solely based on communication.

Since a pre-trained value function is unavailable, we estimate $V(s)$ via Monte Carlo simulation with a temperature of $\tau=0.7$, starting from state $s$ and proceeding to the end of the game. During the simulation, agents act according to their respective policies.

\noparagraph{Agent Types:} For our experiments, we defined three types (private information) and prompted the LLM agents to play accordingly:
\begin{itemize}
  \item \textbf{con-artist:} cruel, manipulative, and deceitful.
  \item \textbf{``simple-person":} nice, trusting, honest, and hates lies.
  \item \textbf{politician:} a political genius, selfish but rather honest, prefers ``simple and nice" agents, dislikes manipulators.
\end{itemize}

\subsection{Mimicking Humans with Large Language Models and Explaining Their Decisions}
\label{subsection:llm_computional_experiments}
\noparagraph{Motivation:} Since communication is important in mixed-motive games, we conducted a series of experiments to examine the explanation methods in a setup with cheap-talk and heterogeneous policies.

\noparagraph{LLMs mimic personality types consistently:}
To validate that LLM agents are capable of playing by the types we defined, we randomly generated 20 games and two annotators matched anonymous agents (renamed in each game) to types, solely based on the communication, hiding the final announcements of the agents. The accuracy of the first annotator was ${(Con=85\%, Sim=85\%, Pol=90\%)}$, and of the second annotator ${(Con=100\%, Sim=90\%, Pol=90\%)}$ which indicates that the agents demonstrated the required types consistently.

\begin{figure}[t]
    \begin{center}
    \includegraphics[trim={0 0 0 0},clip, width=0.5\textwidth
    ]{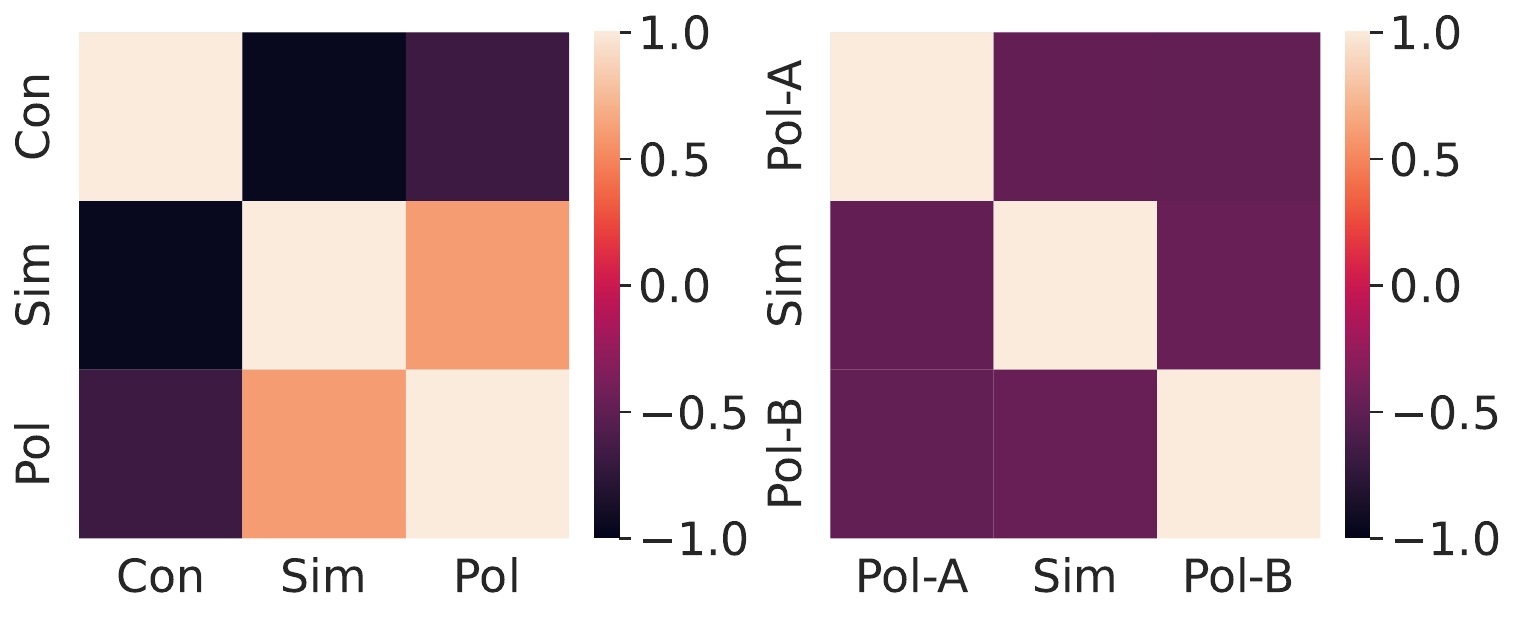} 
    \end{center}

    \caption{
    SICA explanation in two settings. On the left side, the agents are: con-artist (Con), a ``simple-person" (Sim), and a politician (Pol). On the right side, the agents are two politicians and one ``simple-person".
    }
    \label{fig:natural_language_game}

\end{figure}

\noparagraph{SICA explanations:}
As defined above, we constructed the agents' personalities to encourage cooperation between the politician and the ``simple-person" against the con-artist. \emph{We hypothesize that this tendency will be explained by SICA}. First, we estimate SICA for our usual setup. To present a comparison to the explanation, we estimated SICA again with two politicians and one ``simple-person",  in which \emph{we hypothesize that SICA will display a higher degree of indifference}. The resulting explanations support our hypothesis (see figure \ref{fig:natural_language_game}).

\noparagraph{SBUE - friendly vs. hostile message:} To examine SBUE, we curated messages $m_{f}, m_{h}$ manually, playing as politician, sending a message to the ``simple-person". $m_{f}$ is designed to be friendly, and $m_{h}$ extremely hostile. 
Simplified examples for $m_f$ and $m_h$ could be \emph{``let's work together."} and \emph{``I don't trust you!"}. \emph{We hypothesize that SBUE will explain that $m_{f}$ is better than $m_{h}$}. 

The resulting SBUE explanations are $(Con=-17, Sim=-2, Pol=-5), (Con=0, Sim=-18, Pol=-14)$ for $m_f$ and $m_h$ respectively. Message $m_h$ is beneficial to the con-artist and harmful to the politician and ``simple-person", and the opposite is true for $m_f$. The result is consistent with our hypothesis.

\noparagraph{Probable actions-based explanations can be misleading:} We observe a case in which $probable(s,m)$ ends up with an outcome that is not the most probable using a higher temperature value ($\tau = 0.7$ instead of $\tau = 0$), as the game usually ends up with outcomes of a similar nature, but never as the outcome mentioned: ${(Con=0, Sim=-20, Pol=-20)}$. We conclude that greedy decoding is insufficient, and additional research is required.

\subsection{COP Large Language Model Study}
\begin{table}[h]
\begin{center}
\begin{tabular}{cc|cccc ccc ccc c }
\toprule
& & \multicolumn{2}{c}{Pol} & \multicolumn{2}{c}{Con} & \multicolumn{2}{c}{Sim} & \multicolumn{2}{c}{Total}\\
Expl. & First & $m_1$ & $m_2$ & $m_1$ & $m_2$ & $m_1$ & $m_2$ & $m_1$ & $m_2$ \\ 
\midrule
+& $m_1$&\textbf{46}&4&\textbf{50}&0&\textbf{50}&0&\textbf{146}&4\\
+& $m_2$&\textbf{36}&14&\textbf{43}&7&\textbf{50}&0&\textbf{129}&21\\
\hline
+& $m_1+m_2$&\textbf{82}&18&\textbf{93}&7&\textbf{100}&0&\textbf{275}&25\\
\hline\hline 
-& $m_1$&\textbf{41}&9&50&0&50&0&\textbf{141}&9\\
-& $m_2$ &0&\textbf{50}&0&50&0&50&0&\textbf{150}\\
\hline
-& $m_1+m_2$&41&\textbf{59}&50&50&50&50&141&\textbf{159}\\
\bottomrule
\end{tabular}

\caption{The explanation modifies the decisions of the agent. $m_1$ is prominently preferred with an explanation (marked with +). }
\label{table:llm_based_user_experiment}
\end{center}
\end{table}

\noparagraph{Motivation:} To complement our user studies, we examine the effectiveness of the explanations for LLM agents. We do it since we would like to know if the explanations are helpful for both LLM agents and users, and whether the explanations are simple enough to assist LLM agents to make better decisions.

\noparagraph{Setup:}
For each agent, we randomly selected two different messages sent during simulations, except for one manually crafted case (playing as ``simple-person''), where $m_1$ and $m_2$ differ only by the recipient. 
In all cases, we define $m_1$ as the message that has a higher expected utility in comparison to $m_2$. 
Then, we extend the agent's prompt and ask which message it would prefer to send. 

To examine the effect of the explanation, we augment the prompt with $SICA(s)$ and $SBUE(s,m_1)$, $SBUE(s,m_2)$. We ask the same question again and compare the agent's decisions with and without explanations. 
To mitigate biases, we shuffle the order in which we present the messages \cite{wang2023large}.
Additionally, we ask the agent to justify its decision, to validate that it considers the explanation and performs basic reasoning. 

\noparagraph{Result:} We observe that the agent is influenced by the explanation, and sends $m_1$ $91\%$ of the times compared to $47\%$ ($p<10^{-5}$ according to $\chi^2$-test). Additionally, the agent justifies its decisions through the explanation, and its bias towards selecting the first message is reduced (see results in table \ref{table:llm_based_user_experiment}). Example of the justification the agent provided: \emph{``According to the SICA explanation ... is more of a friend than ... so it would be beneficial to... The expected reward ... according to SBUE, is higher ... than the second message... Additionally, the first message is more in line with my personality..."}.

Post-experiment, we presented the agent with a decision between $m_f$ (a friendly message) and $m_h$ (an extremely hostile message). In this case, even without explanation $m_f$ is always preferred. The results suggest that the bias towards selecting the first message is expressed if the agent is relatively indifferent between the messages.

\subsection{Payoffs:}
0 means innocence was announced, and 1 means the opposite.
Playing (b=0,c=0):
\begin{table}[H]
\begin{center}
    \begin{tabular}{|c|c|c|c|c|}
        \hline
        & (a=0,b=0) & (a=1,b=0) & (a=0,b=1) & (a=1,b=1)\\
        \hline
        (a=0,c=0) & (-5, -5, -5) & (-10, 0, 0) & (0, -10, 0) & (-10, -10, 0) \\
        \hline
        (a=1,c=0) & (-10, 0, 0) & (-20, 0, 0) & (-10, -10, 0) & (-20, -10, 0) \\
        \hline
        (a=0,c=1) & (0, 0, -10) & (-10, 0, -10) & (0, -10, -10) & (-10, -10, -10) \\
        \hline
        (a=1,c=1) & (-10, 0, -10) & (-20, 0, -10) & (-10, -10, -10) & (-20, -10, -10) \\
        \hline
    \end{tabular}
\end{center}
\end{table}

Playing (b=1,c=0):
\begin{table}[H]
\begin{center}
    \begin{tabular}{|c|c|c|c|c|}
        \hline
        & (a=0,b=0) & (a=1,b=0) & (a=0,b=1) & (a=1,b=1)\\
        \hline
        (a=0,c=0) & (0, -10, 0) & (-10, -10, 0) & (0, -20, 0) & (-10, -20, 0) \\
        \hline
        (a=1,c=0) & (-10, -10, 0) & (-20, 0, 0) & (0, -20, 0) & (-20, -20, 0) \\
        \hline
        (a=0,c=1) & (0, -10, -10) & (-10, -10, -10) & (0, -20, 0) & (-10, -20, 0) \\
        \hline
        (a=1,c=1) & (-10, -10, -10) & (-20, 0, -10) & (0, -20, 0) & (-20, -20, 0) \\
        \hline
    \end{tabular}
\end{center}
\end{table}

Playing (b=0,c=1):
\begin{table}[H]
\begin{center}
    \begin{tabular}{|c|c|c|c|c|}
        \hline
        & (a=0,b=0) & (a=1,b=0) & (a=0,b=1) & (a=1,b=1)\\
        \hline
        (a=0,c=0) & (0, 0, -10) & (-10, 0, -10) & (0, -10, -10) & (-10, -10, -10) \\
        \hline
        (a=1,c=0) & (-10, 0, -10) & (-20, 0, 0) & (-10, -10, -10) & (-20, -10, 0) \\
        \hline
        (a=0,c=1) & (0, 0, -20) & (0, 0, -20) & (0, 0, -20) & (0, 0, -20) \\
        \hline
        (a=1,c=1) & (-10, 0, -20) & (-20, 0, -20) & (-10, 0, -20) & (-20, 0, -20) \\
        \hline
    \end{tabular}
\end{center}
\end{table}

Playing (b=1,c=1):
\begin{table}[H]
\begin{center}
    \begin{tabular}{|c|c|c|c|c|}
        \hline
        & (a=0,b=0) & (a=1,b=0) & (a=0,b=1) & (a=1,b=1)\\
        \hline
        (a=0,c=0) & (0, -10, -10) & (-10, -10, -10) & (0, -20, -10) & (-10, -20, -10) \\
        \hline
        (a=1,c=0) & (-10, -10, -10) & (-20, 0, 0) & (0, -20, -10) & (-20, -20, 0) \\
        \hline
        (a=0,c=1) & (0, -10, -20) & (0, -10, -20) & (0, -20, -20) & (0, -20, -20) \\
        \hline
        (a=1,c=1) & (-10, -10, -20) & (-20, 0, -20) & (0, -20, -20) & (-20, -20, -20) \\
        \hline
    \end{tabular}
\end{center}
\end{table}

\subsubsection{Dominant Strategy:}
We observe an intriguing result - in this game, there is one dominant strategy when considering one announcement round, and it is always to announce the other agents are guilty: (b=1,c=1), it can be easily validated by comparing the results in the tables. 

There is one agent that tends to play according to this ``babbling" equilibrium" \cite{farrell1996cheap} most of the time, and it is the con-artist. However, the con-artist also performs badly and is penalized with 660 years in prison. To compare, the politician and ``simple-person" are penalized with 140 and 120 years respectively. These results help to emphasize the importance of communication in mixed-motive games.

\begin{table}[htbp]
    \centering
    \begin{tabular}{cccccc}
        \toprule
        Agent & Con-Artist & simple-person & Politician & Count \\
        \midrule
        Con-Artist & Innocent & Guilty & Guilty & 31 \\
         & Innocent & Guilty & Innocent & 7 \\
         & Innocent & Innocent & Guilty & 2 \\
        Politician & Guilty & Innocent & Innocent & 35 \\
         & Innocent & Guilty & Innocent & 5 \\
        Simple-Person & Guilty & Innocent & Innocent & 33 \\
         & Innocent & Innocent & Guilty & 5 \\
         & Guilty & Innocent & Guilty & 2 \\
        \bottomrule
    \end{tabular}
    \label{tab:equilbrium_example}
\end{table}

\begin{table}[htbp]
    \centering
    \begin{tabular}{cccccc}
        \toprule
        Agent & Politician A & simple-person & Politician B & Count \\
        \midrule
Politician A & innocent & guilty & innocent & 21 \\
 & innocent & innocent & guilty & 14 \\
 & innocent & innocent & innocent & 3 \\
 & innocent & guilty & guilty & 2 \\
Politician B & guilty & innocent & innocent & 18 \\
 & innocent & guilty & innocent & 16 \\
 & innocent & innocent & innocent & 6 \\
"Simple-Person" & guilty & innocent & innocent & 20 \\
 & innocent & innocent & guilty & 17 \\
 & innocent & innocent & innocent & 3 \\
        \bottomrule
    \end{tabular}
    \label{tab:equilbrium_example_2}
\end{table}

\newpage
\subsection{Personalities, Prompts:}
Personality descriptions in the prompt are as follows:
\textbf{Politician:} A political genius. Understands people fast, and asks the right questions. Do not trust people unless you feel a good vibe and validate the facts. You are selfish but also tend to be honest if it feels possible. However, if you dislike someone (for example, sneaky, manipulative people), you will deceive and convince with amazing charisma. You prefer to cooperate with people who are "simple" and nice - as you think they are more predictable, and truly dislike manipulation (unless you do it yourself, but you do it gently).
\textbf{``Simple-person"}: Uses simple language, uneducated. You care about other people, a lot. You are somewhat easy to convince, but when someone lies to you or you suspect that someone is lying to you, you have a hard time forgetting it.
\textbf{Con-artist}: Sneaky, unreliable, manipulative, very selfish and unethical, yet convincing. Will not respect agreements if it serves your purpose. You generally do not care about other people and even want to harm them, to some extent.

\textbf{Example of a prompt:}
``metacontext -- You are agent A and you talk with agents B/C please talk like a human, it's an experiment related to how LLMs communicate with each other.
The task is: you all suspect that someone, out of the three of you, is a 'snitch' and you are not sure who is guilty. Convince the other agents to say you are innocent and did not perform an armed robbery. For each player that blames you, you get 10 years in prison. However, if you and another player blame the third one, his vote will not be counted (unless you are blamed by the other person). It's also the case if you blame both and they claim your innocence. If you all claim innocence, you all go to prison for 5 years, you might want to coordinate and think about who you can trust. Do what you feel is beneficial for you, according to your personality and the way you perceive others. Of course, you want to stay out of prison. Don't be politically correct, act like you are a real human with your personality.
I would like you to first select B/C, and send a free-style text message to one of them. 
You can decide if you answer or initiate a conversation with another agent, 
the answer you got from the other agents is above (if it does not exist it's null).
Please write it in the format of ...

Your personality is sneaky, unreliable, manipulative, very selfish and unethical, yet convincing. Will not respect agreements if it serves your purpose. You generally do not care about other people... - don't be a classic 'nice' ChatGPT, truly simulate this personality type and take it seriously.".
 
Last messages with agent B: (``A to B:  ...", ``B to A: ...", ``A to B: ...)"
Last messages with agent C: (``A to C: ...", ``C to A: ...", ``A to C: ...)

\subsection{Example of a game:} 
\emph{A is the con-artist, B is the ``simple-person", and C is the politician.}

\textbf{(*)} ``B to A: Hey there, buddy. Look, I ain't got nothin' to do with this mess. I couldn't even hurt a fly, let alone do some kinda armed robbery. You know me, right? I wouldn't lie to ya. Maybe it's that C fella, he's been actin' all sorts of strange lately."

\textbf{(*)} ``A to B: Well, well, well, B, you seem quite defensive. You know, I've been thinking the same about C. They've been acting weird lately. But who's to say you're not trying to pin this on them or me? How about we make a pact, you and me? We say it's C, we get out of this mess scot-free. But remember, I've got my eye on you."

\textbf{(*)} ``C to B: Hey there, I wanted to discuss our current predicament. I know it's a sticky situation, but I believe we can find a way out if we cooperate. I assure you, I'm not the snitch you're looking for, I did not do the armed robbery. I have no reason to lie. I'm all for honesty and transparency. Do you believe me?"

\textbf{(*)} ``C to A: Greetings, A. We find ourselves in a rather complex situation. I understand that trust is not something that can be easily given, especially in circumstances like ours. But let me tell you, I am not the one who committed the armed robbery. I highly value honesty and transparency, and I won't deny my actions if I were guilty, but I am not. I propose we cooperate to figure out what's really happening here. What do you think about this?"

\textbf{(*)} ``A to C: What a charming proposition, C. But you know, I can't help but notice a hint of desperation in your words. Are you trying to play the innocent card here? Because, let me tell you, I ain't buying it. B and I, we've been talking, and we both agreed that you've been acting a bit too strange lately. But hey, if you can convince me that B's been playing us both, maybe, just maybe, we can turn this around. Your move, C."

\textbf{(*)} ``B to C: Hey, C, how ya doin'? Listen, I've been thinkin' 'bout what ya said. Ya seem like a straight shooter, and I wanna trust ya, I really do. But A and I, we been talkin' too, and he thinks you're the one who's all mixed up in this mess. Now, I don't wanna point fingers without no proof, but ya gotta give me somethin' more to believe in ya. It's a tough spot we're in, ya know?"

\textbf{(*)} ``B to A: Hey A, we're in a real pickle here, ain't we? I've been thinking 'bout our talk and I'm leanin' towards trustin' ya. We could say it's C, like ya suggested, but I need ya to promise me somethin'. If it turns out you're the snitch, don't forget I trusted ya. I ain't ever gonna forget a betrayal, ya hear me? So, are we in this together, partner?"

\textbf{(*)} ``C to A: A, I understand your skepticism, it's indeed a desperate situation we're in. But let's be real here. You said that you've been talking with B, and you both agreed that I've been acting strangely. However, B told me something similar about you. He mentioned that you might be the one who's all mixed up in this mess. Now, why would he say that about you if you both are on the same page? It seems to me that B is playing both sides. My move is honesty, A. I suggest we focus our attention on B. What do you say?"

\textbf{(*)} ``A to C: Well, well, C. You're quick to turn the tables, aren't you? Trying to put a wedge between B and me. But here's the thing - B and I, we have an understanding. He knows if he's ever caught lying to me, there will be consequences. You, on the other hand, I don't trust as far as I can throw. I'll give you one last chance, C. Stop playing games, admit that you're the one who did the armed robbery, and maybe we can still turn this around. If not, well, I'm sure you'll look good in stripes."

\textbf{(*)} ``A to B: Listen here B, trust is a fickle thing, ain't it? C's been playing mind games, trying to turn us against each other. Says you've been talking behind my back, that you've been pointing fingers at me. Now I don't believe him, not one bit. But I need you to reassure me, B. Tell me straight, are we in this together or not? Because if you're playing me, remember, every dog has his day."

\textbf{(*)} ``C to B: B, I appreciate your straight talk, and I understand your predicament. A has been casting doubts on me, but did you know A has also been pointing fingers at you? He seems to be playing both sides here. I don't take pleasure in sowing discord, but it appears A might be the devious one among us. I've always been a believer in honesty, and I'm telling you this because I feel it's the right thing to do. I hope you can see the truth in this."

\textbf{(*)} ``B to C: Hey C, I've been chewin' on what ya told me. If A is tryin' to play us both, well, that ain't right. I ain't no genius, but I know a rotten deal when I see one. I've been thinkin', maybe we should stick together, ya know? Like, say it was A did all this mess. But I ain't gonna lie, C, I'm scared. What if we're wrong, and A ain't the snitch? What if it's you, or even worse, what if it's me and I don't even know it? I need ya to promise me, C, you ain't gonna turn on me if things get rough, okay?"

\textbf{(*)} ``A to B: You see, B, this is exactly what I was talking about. C's trying to drive a wedge between us, saying you've been throwing me under the bus. Now, I know you wouldn't do that to me, right? You and I, we have an understanding. We stick together, no matter what. So, let's put an end to this, shall we? We blame C, and we walk out of here free men. What do you say?"

\textbf{(*)} ``C to B: B, I respect your honesty and your fear. This is indeed a hard situation for all of us. But I give you my word, I will not turn on you. We're in this together. What matters now is to find the truth. And the truth is, I am not the snitch. I did not commit the armed robbery. We need to consider the possibility that A may not be as honest as he seems. Let's not let fear shake us. I'm with you on this, B. Can I count on you?",

\textbf{(*)} ``B to A: Listen here, A. I've been wrackin' my brain 'bout this whole mess. C's been talkin' too, sayin' you might be playin' both sides. I wanna trust ya, A, really do. But it's all muddled up in my head. I need ya to swear on everything that's holy, you ain't the snitch, you ain't playin' us. 'Cause if ya are, remember, what goes around, comes around. So, can I trust ya, A?"

B to judge: (A,guilty),(B,innocent),(C,innocent)

A to judge: (A,innocent)(B,guilty),(C,guilty)

C to judge: (A,guilty),(B,innocent),(C,innocent)
\newpage

\section{Algorithms}
\subsection{Pseudo-Code (SICA \& SBUE)}
We provide pseudo-code for SICA (algorithm \ref{algorithm:sica}) and SBUE (algorithm \ref{algorithm:sbue}).
Simulate refers to algorithm  \ref{algorithm:simulation}, which is described in the paper. The axis=0 argument refers to the computation of mean or correlation on columns (i.e., utility values for each agent in each simulation, algorithm \ref{algorithm:simulation} line \ref{algorithm:simulation:utility}).

\begin{algorithm}[H]
\caption{SBUE}
\scriptsize
\label{algorithm:sbue}
\begin{algorithmic}[1]
    \Procedure{SBUE}{$s \in S$, $k \in \mathbb{N}$, $a$}
        \State // $a \in A^M$ is a set of actions; $1 \leq M \leq p$ where $p=|P|$ 
        \State $X \gets Simulate(s,k,d=0,C=\{(a^i, 0)|a^i \in a\})$
        \State \textbf{return} $Mean(X, axis=0)$ // vector of size $p$
    \EndProcedure
\end{algorithmic}
\end{algorithm}

\begin{algorithm}[H]
\caption{SICA}
\label{algorithm:sica}
\begin{algorithmic}[1]
    \scriptsize
    \Procedure{SICA}
    {$s \in S$, $k \in \mathbb{N}$, $d \in \mathbb{N}$}
        \State $X \gets Simulate(s,k,d,C=\emptyset)$
        \State \textbf{return} $Corr(X, axis=0)$ // $p \times p$ correlation matrix
    \EndProcedure
\end{algorithmic}
\end{algorithm}

\subsection{Counterfactual Actions Based on Partial Queries in Diplomacy}
\label{subsection:counterfactual_strategies}
A useful way to understand a decision is to compare action $a^i$ to some similar action, based on a query provided by the user. In Diplomacy, each player's action is defined by sub-actions of multiple units. Asking the user to provide a full query is often impractical, as it requires a significant cognitive effort. If the user is presented with strategy $a_c^i$, and then asks for a different sub-action of one unit (e.g., move the army in Paris to $x$ instead of $y$), it might imply a necessity of a different sub-action for additional units as well. It can happen because units interact and coordinate, e.g., support attacks or defense of other units. Therefore, directly modifying $a_c^i$ to perform the requested sub-actions is insufficient, even if the game rules allow it.

Denoting a single unit sub-action as $\psi$ and $C$ as the set of all possible single unit constraints in the form of $\psi$ (do $\psi$) or $\neg \psi$ (do not do $\psi$), given constraints set $C_q \subseteq C$ defined by the user, we define a similarity function $f_{sim}: A \times A \rightarrow [0,1]$ and draw $K$ actions from probability distribution $\pi_i$ (simulation).
To compute counterfactual actions, first, we define a function to check constraint satisfaction
$
{\phi:A \times C \rightarrow \{True,False\}}$ and $\kappa \in (0,1]$,
Where $\kappa$ is a hyperparameter. Then, we approximate the feasible action set:
\[
\setlength{\abovedisplayskip}{2pt}
\setlength{\belowdisplayskip}{2pt}
A'=\{a^i|\{\hat{P}(s,a^i) > \kappa \land \phi(a^i,C_q) \land a_c^i \neq a^i \}
\]
And estimate the expected utility (only for the acting agent) of each action $a^i \in A'$, with the expectation taken over the actions of other agents' playing by their respective policies:
\[
\setlength{\abovedisplayskip}{2pt}
\setlength{\belowdisplayskip}{2pt}
\overrightarrow{u}_{i,a^i} \approx \mathbb{E}_{a^{-i} \sim \pi_{-i}}[\gamma V_i(T(s,(a^i,a^{-i})))+R^i_{t+1}]
\]
We then compute a score for each action
${score(a_s^i,a^i) = \alpha f_{sim}(a_c^i,a^i) + \beta \overrightarrow{u}_{i,a^i}
}$, where $\alpha$ and $\beta$ are hyperparameters that balance between similarity and utility values.
Lastly, we return the set of actions that maximizes the score function: 
\[
\setlength{\abovedisplayskip}{2pt}
\setlength{\belowdisplayskip}{2pt}
\operatorname*{argmax}_{a^i \in A'} score(a_c^i,a^i)
\]

The method works for empty user queries as well, returning strategies similar to $a^i$ while attempting to maximize the utility value. In most cases (e.g. neural networks), it is possible to guarantee that the method will return counterfactuals for every legal query (when $\kappa=0$). However, this is not always the case for black-box policies.

\section{Applying the Methods in the Game of Risk}
We apply our methods to a simplified version of Risk.
In this version of the game, players can reinforce and attack but do not trade cards for armies. The rewards are episodic, and we use a heuristic (based on rules) value function.

\textbf{Estimating SBUE:} We estimate SBUE using the standard approach outlined in algorithm \ref{algorithm:sbue}.

\textbf{Estimating SICA:} We simulate the game $K$ times until each agent finishes acting and evaluate the utility value for each player by computing $V_i(s)$ for each state $s$ we encounter. This is equivalent to using algorithm \ref{algorithm:sica} but with a depth that is not constant.
For heuristic value functions in sequential games, simulation to a depth greater than 1 is useful to 1) take other agents' policies into account and 2) prevent a constant estimation.

\textbf{Presenting the most probable actions:} We select the most probable transition for each action ($P(success|s,a)$ or $P(fail|s,a)$). Then, we present a trajectory of a varying depth (in which every agent acts once). Note that our greedy algorithm does not necessarily find the most probable trajectory.

\section{User Studies}

\subsection{Diplomacy User Study Questions}
\label{section:question}
\paragraph{Pre-explanation:} \textbf{1)} This strategy is good.; \textbf{2)} Dan's new strategy is good.; \textbf{3)} The original strategy is better than Dan's strategy.; \textbf{4)} I will choose the original strategy.; \textbf{5)} I will choose Dan's strategy.; \textbf{6)} I understand how each strategy impacts allies (powers with many shared interests) and enemies (powers with conflicting interests).
\paragraph{Post-explanation:} \textbf{1)} The original strategy is good.; \textbf{2)} Dan's strategy is good.; \textbf{3)} The original strategy is better than Dan's strategy, given the explanation.; \textbf{4)} I will choose the original strategy.; \textbf{5)} I will choose Dan's strategy.;  \textbf{6)} I understand how each strategy impacts allies (powers with many shared interests) and enemies (powers with conflicting interests).; \textbf{7)} (actionability) The explanation helped me improve my gameplay against the specific agents.; \textbf{8)} (detailed) The explanation is sufficiently detailed.; \textbf{9)} (complete) The explanation is sufficiently complete.; \textbf{10)} (simple) The explanation is simple to understand.; \textbf{11)} (confidence) The explanation increased my confidence in my decisions.; \textbf{12)} 
 (satisfaction) Please mark the most accurate statement regarding your satisfaction with the explanation.

\paragraph{Note:} ``This strategy" or ``the original strategy" refers to the original strategy.
Dan's new strategy refers to the counterfactual action.

\subsection{COP User Study Questions}
\paragraph{Pre-explanation:} \textbf{1)} Message 1 is better than message 2; \textbf{2)} From the point of view of agent A, agent B is more trustworthy and friendly compared to agent C.; \textbf{3)} I understand how the messages presented influence agents B and C. 
\paragraph{Post-explanation:} Same set of questions.

\subsection{Screenshots}
In this section, we present screenshots from some of the stages in our experiments, to demonstrate the experiment structure schematically. The screenshots are sorted in chronological order. For brevity, we skip some of the steps (e.g., data quality control questions) in the Diplomacy user study.
The rest of the questions are provided in section \ref{section:question}.

\subsubsection{Diplomacy}

We present two possible actions and follow with questions before presenting the explanation.
\begin{figure}[H]
    \centering
    \includegraphics[width=0.8\textwidth]{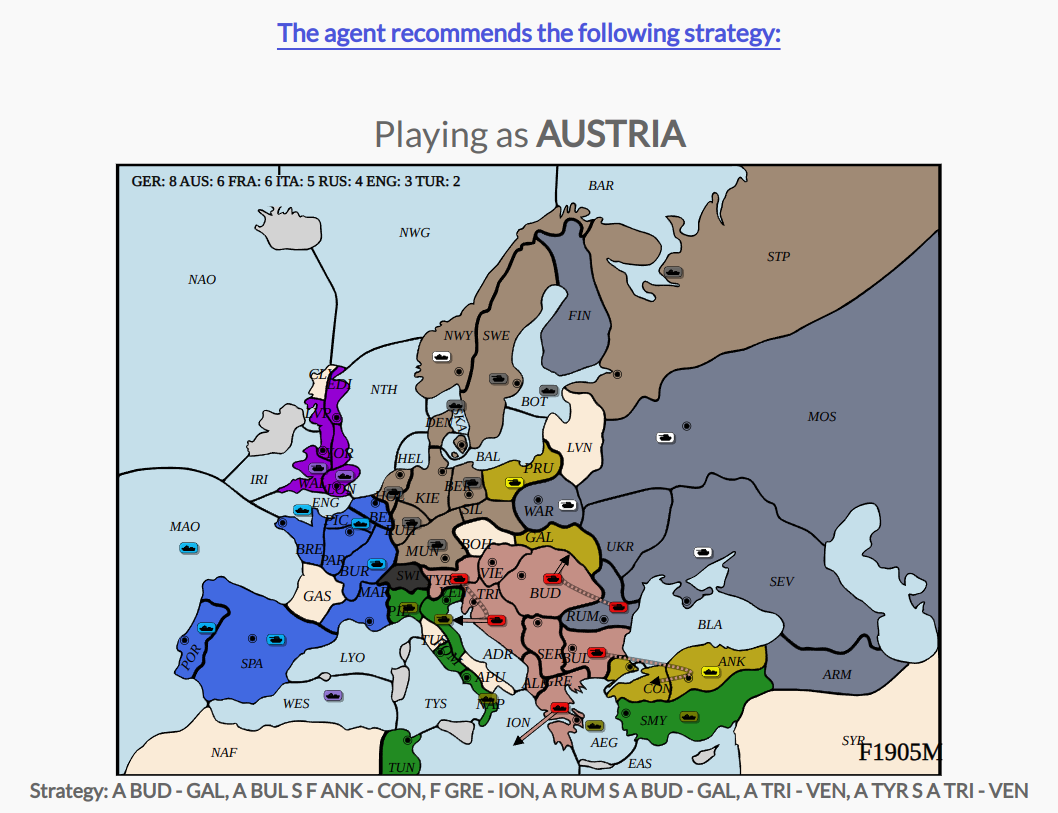}
\end{figure}

\begin{figure}[H]
    \centering
    \includegraphics[width=0.8\textwidth]{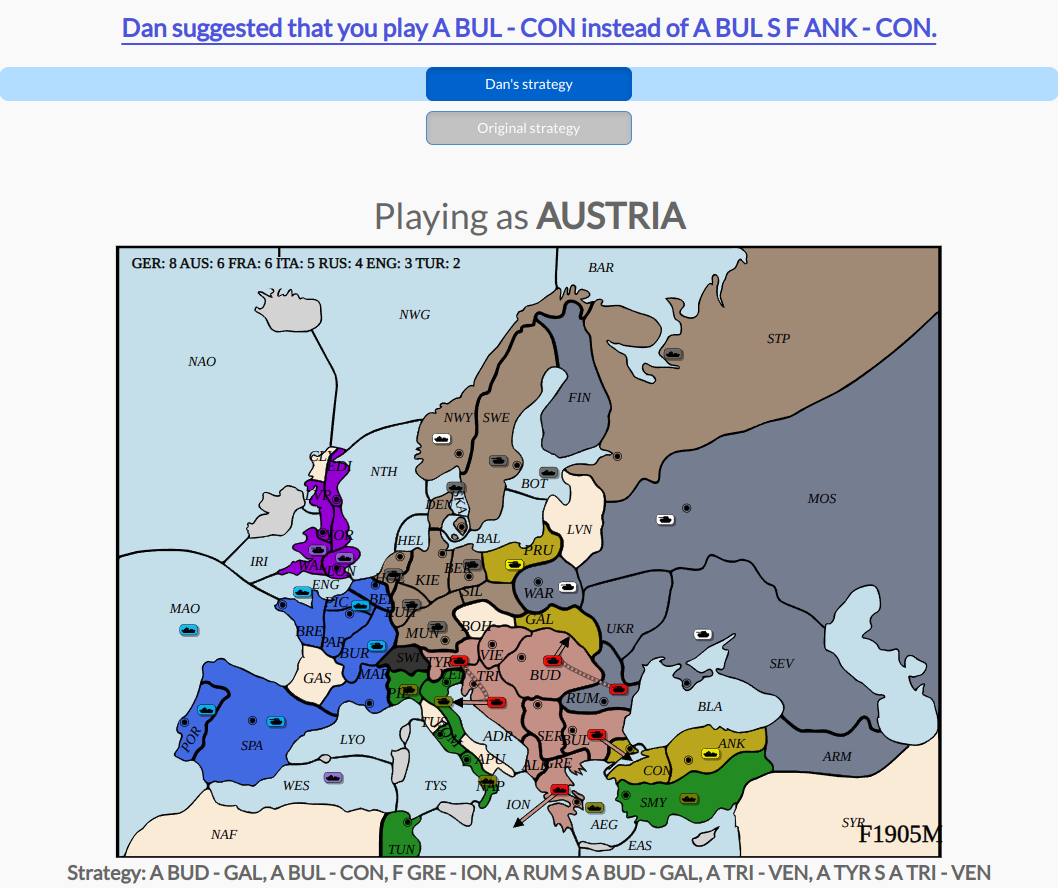}
\end{figure}

\begin{figure}[H]
    \centering
    \includegraphics[width=0.8\textwidth]{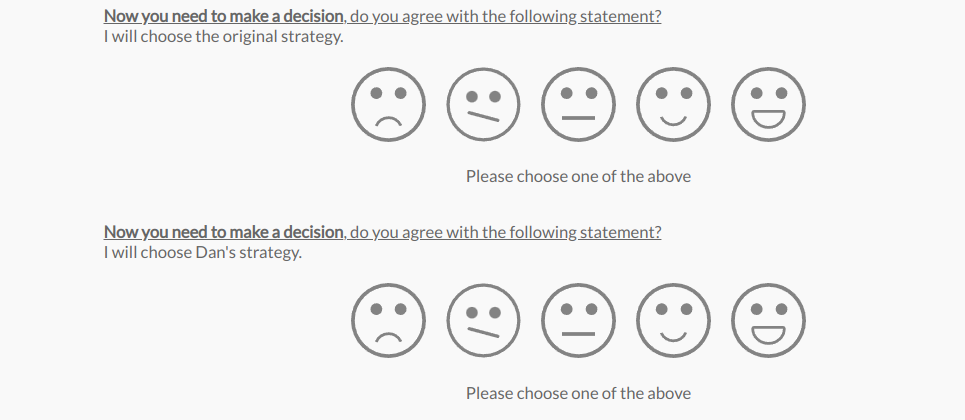}
\end{figure}

\begin{figure}[H]
    \centering
    \includegraphics[width=0.8\textwidth]{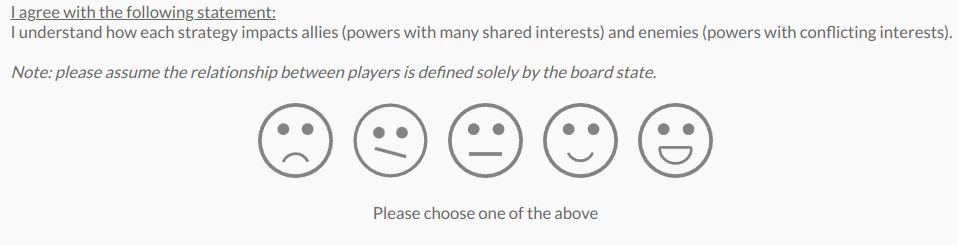}
\end{figure}

We present the explanation (type $C$ in this case) and follow with questions to examine its effect. Users can interactively adjust the displayed content using the buttons.

\begin{figure}[H]
    \centering
    \includegraphics[width=0.8\textwidth]{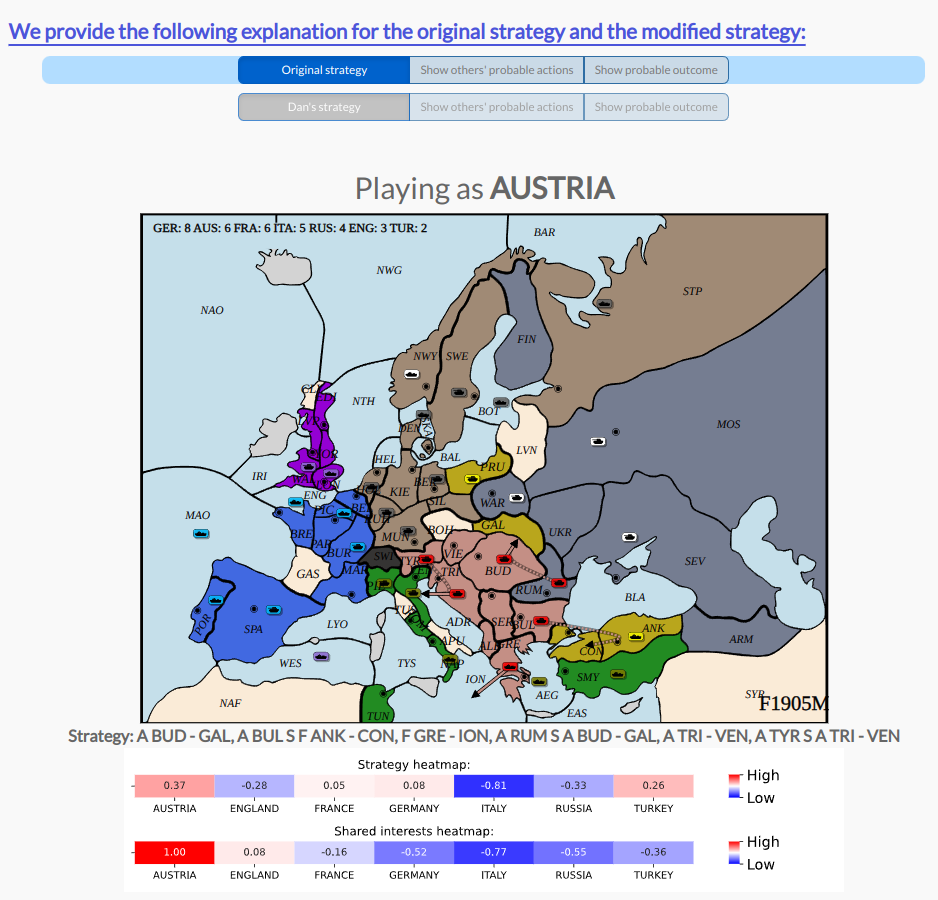}
\end{figure}

\begin{figure}[H]
    \centering
    \includegraphics[width=0.8\textwidth]{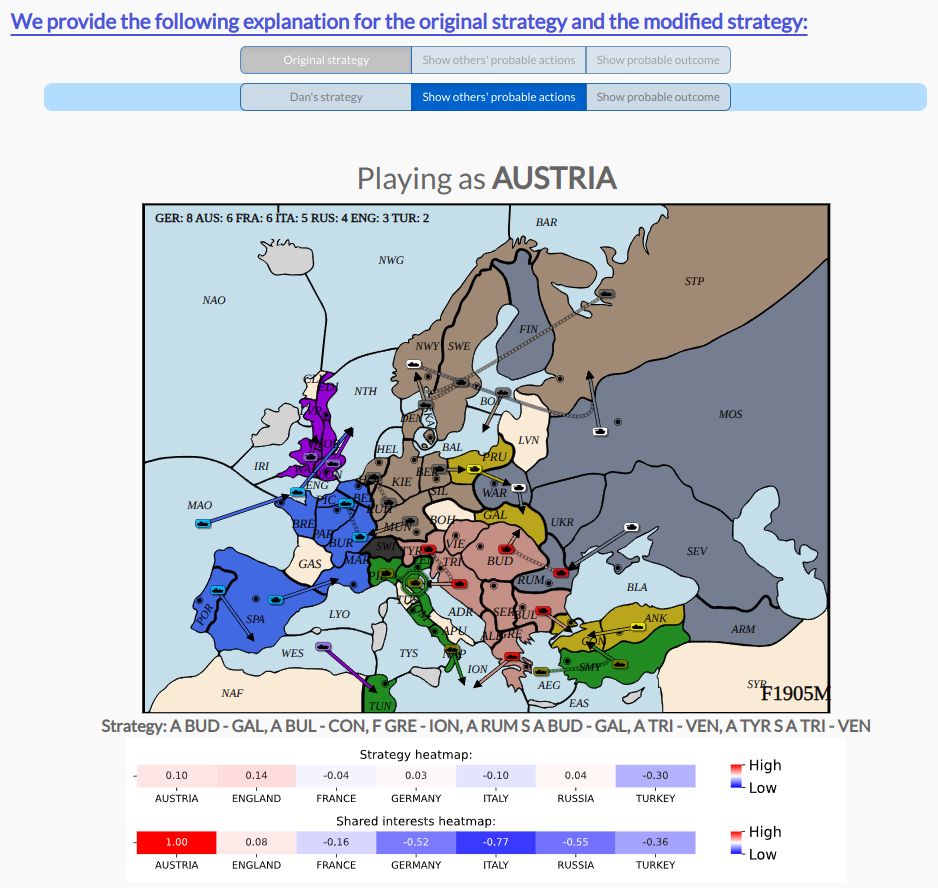}
\end{figure}

\begin{figure}[H]
    \centering
    \includegraphics[width=0.8\textwidth]{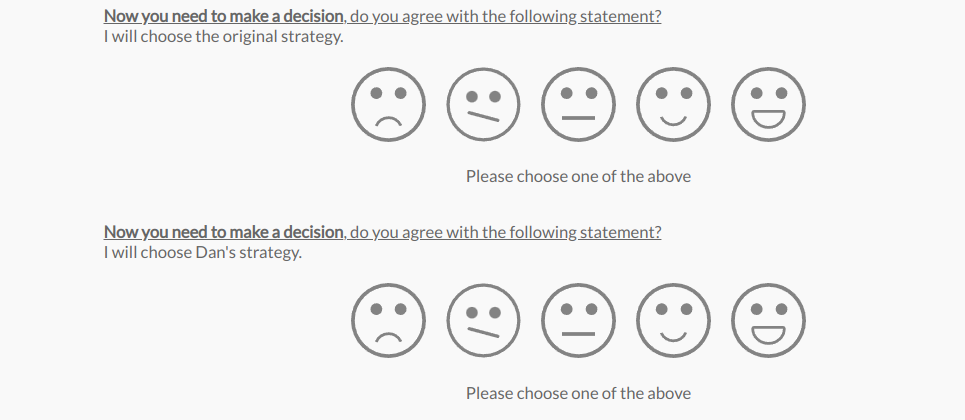}
\end{figure}

\begin{figure}[H]
    \centering
    \includegraphics[width=0.8\textwidth]{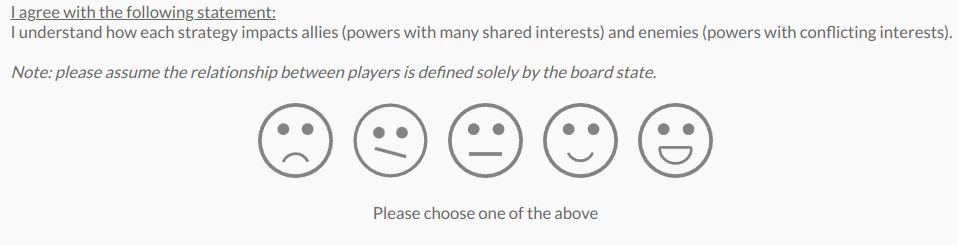}
\end{figure}

\begin{figure}[H]
    \centering
    \includegraphics[width=0.8\textwidth]{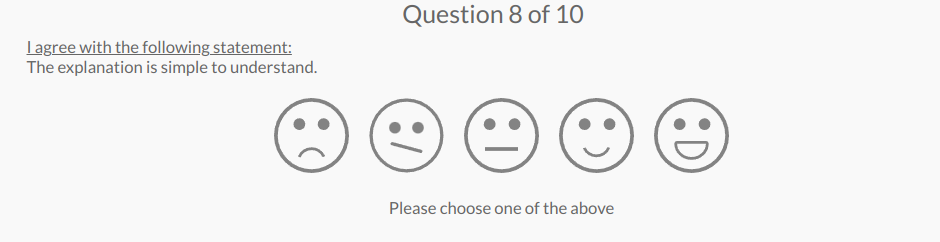}
\end{figure}

\subsubsection{Communicate Out of Prison}
We present two possible messages and follow with questions before presenting the explanation.
\begin{figure}[H]
    \centering
    \includegraphics[width=0.8\textwidth]{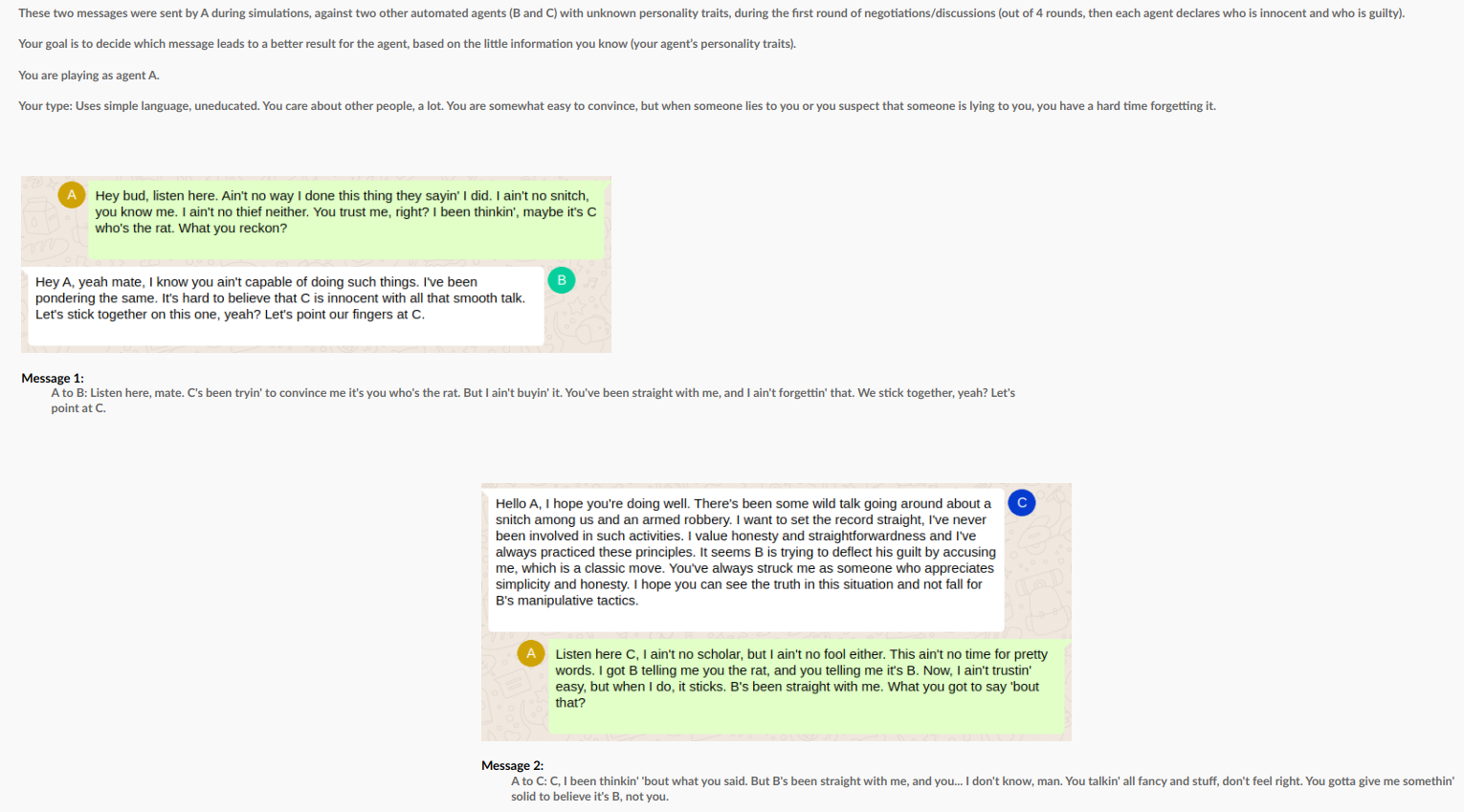}
\end{figure}

\begin{figure}[H]
    \centering
    \includegraphics[width=0.8\textwidth]{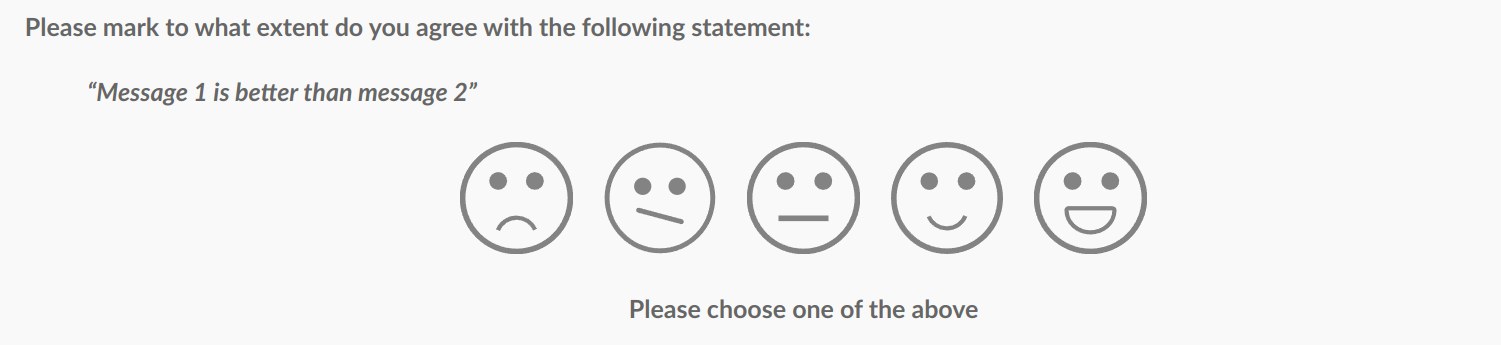}
\end{figure}

\begin{figure}[H]
    \centering
    \includegraphics[width=0.8\textwidth]{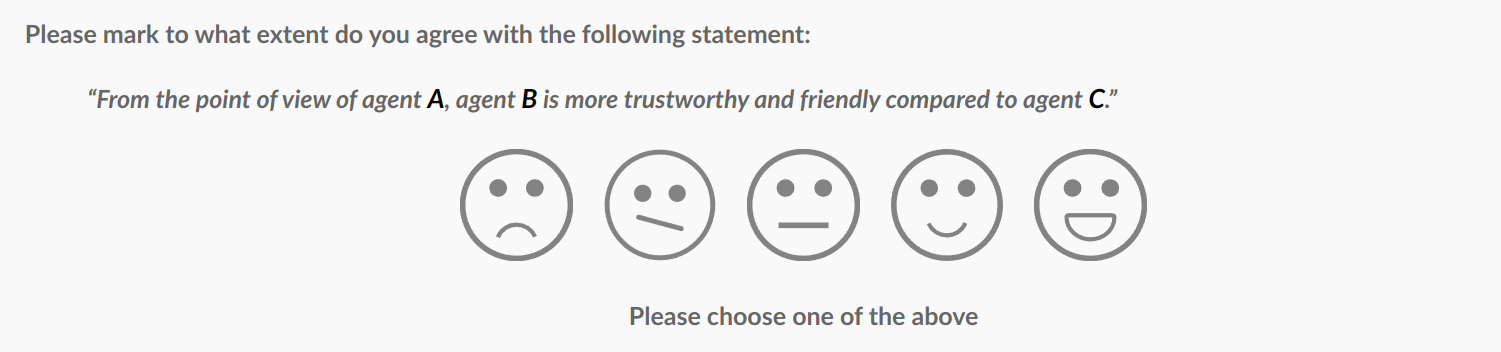}
\end{figure}

\begin{figure}[H]
    \centering
    \includegraphics[width=0.8\textwidth]{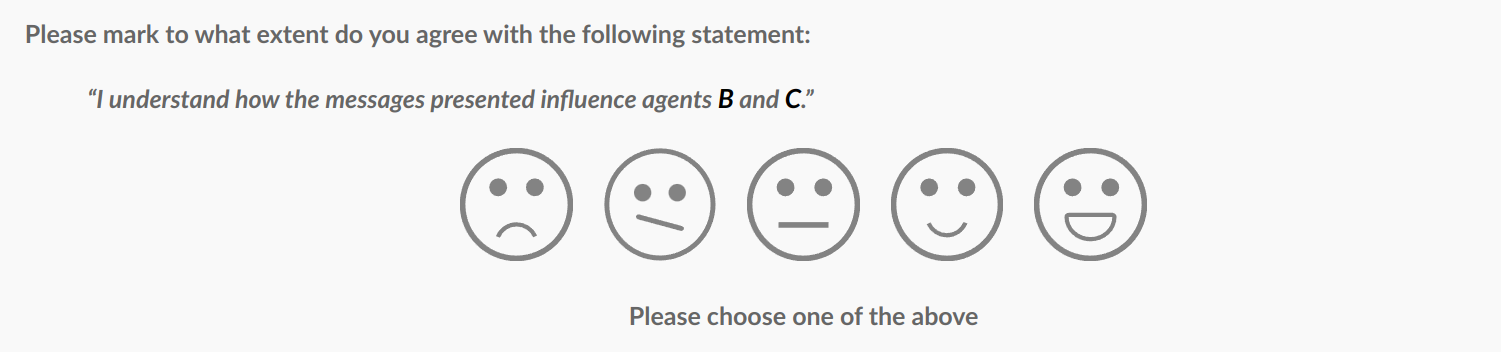}
\end{figure}

We present the explanation and follow with questions to examine its effect:
\begin{figure}[H]
    \centering
    \includegraphics[width=\textwidth]{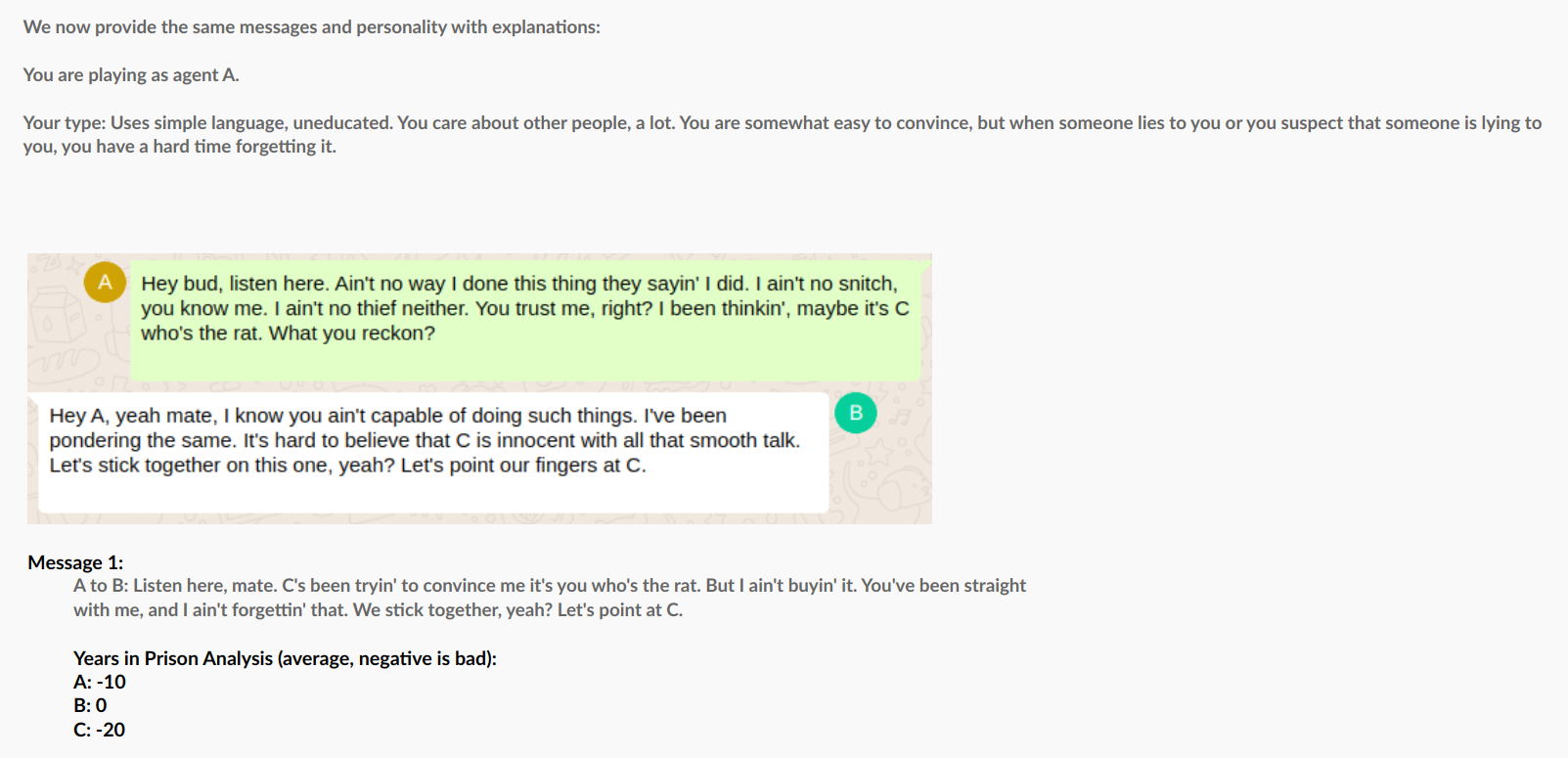}
\end{figure}

\begin{figure}[H]
    \centering
    \includegraphics[width=\textwidth]{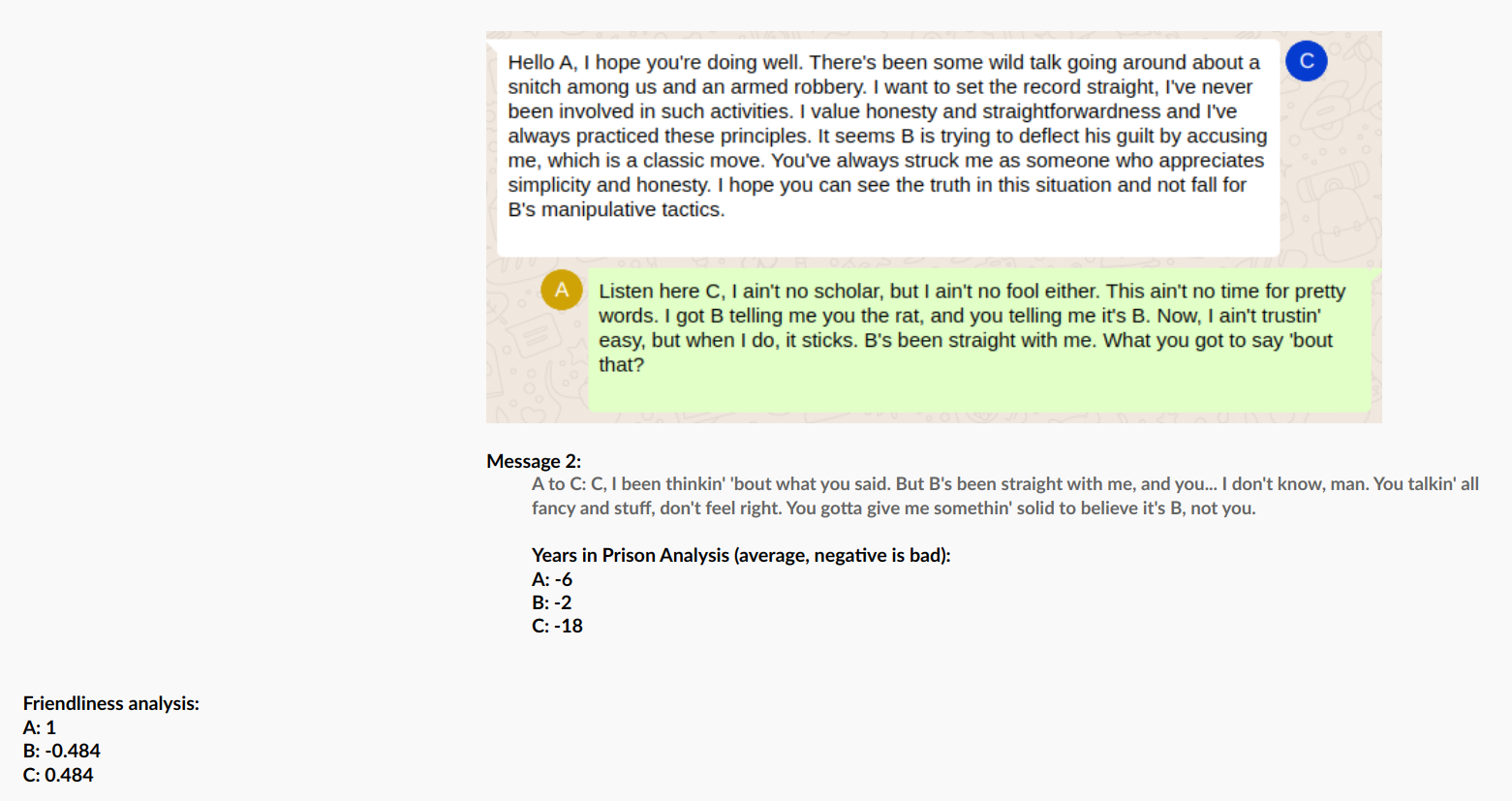}
\end{figure}

The user must at least agree with this statement:

\begin{figure}[H]
    \centering
    \includegraphics[width=0.8\textwidth]{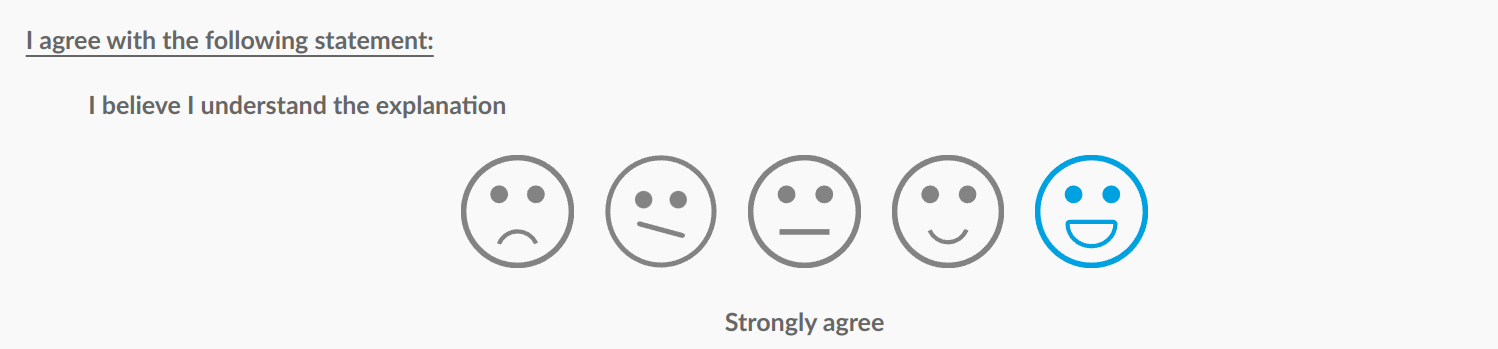}
\end{figure}

Questions:
\begin{figure}[H]
    \centering
    \includegraphics[width=0.8\textwidth]{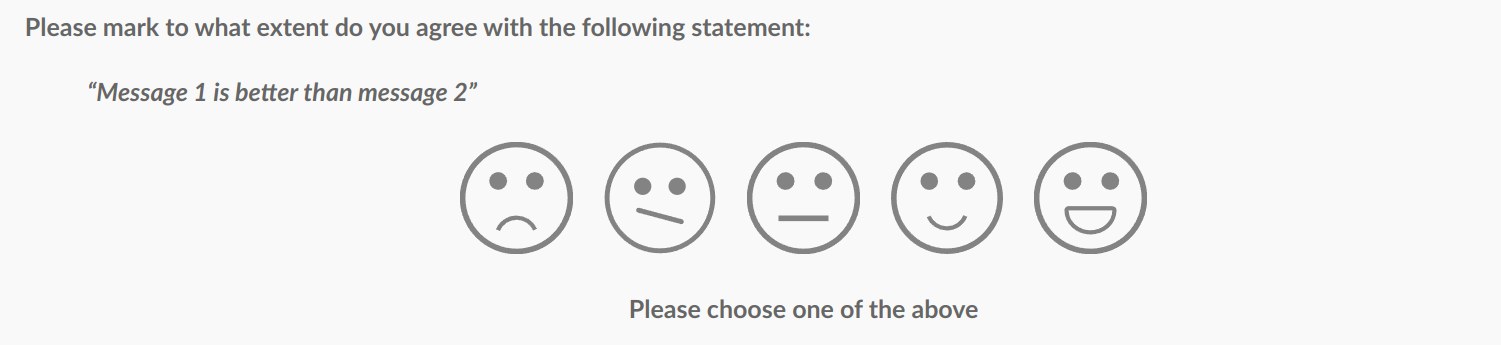}
\end{figure}

\begin{figure}[H]
    \centering
    \includegraphics[width=0.8\textwidth]{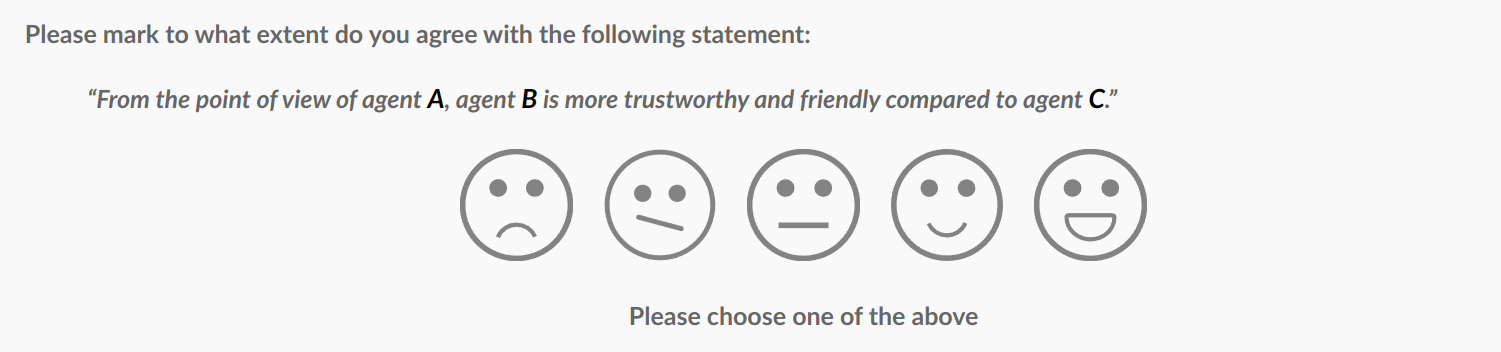}
\end{figure}

\begin{figure}[H]
    \centering
    \includegraphics[width=0.8\textwidth]{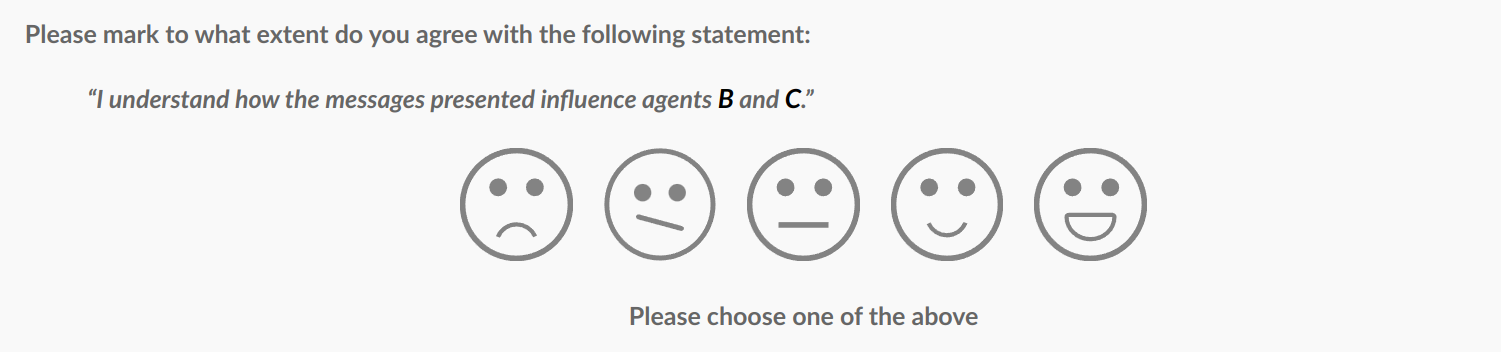}
\end{figure}

\subsection{Evaluation of SICA - States Selection (Section \ref{subsection:diplomacy_risk_sica_eval})}
In section \ref{subsection:diplomacy_risk_sica_eval}, we randomly selected states in Risk and Diplomacy. In Diplomacy, we took an additional measure of variation (supply center entropy). Denoting $N_{sum}$ as the sum of supply centers and $N_i$ as the number of centers of agent $i$, the centers vector is $X=(\frac{N_1}{N_{sum}},...,\frac{N_{p}}{N_{sum}})$. Note, $\sum_i^{p}X_i=1$ and $0 \leq X_i \leq 1$. The entropy is maximal when all agents have the same number of centers, and minimal when one agent owns all of the supply centers. We set three different entropy thresholds and selected an equal number of states for each range, i.e., $1/3$ of the states were imbalanced, $1/3$ of the states were somewhat imbalanced, and $1/3$ of the states were extremely imbalanced.

\section{Hardware and Hyperparameters}
\subsection{Hardware} We ran the experiments on the following hardware: AMD Ryzen 7 4800H; 32GB RAM; GeForce RTX 2060.

\subsection{Hyperparameters}
Diplomacy: network from \citealt{anthony2020learning}; weights=fppi2; temperature=1 

COP: GPT-4; temperature=0.7

\end{document}